%% file: transistorformer-17.tex
\documentclass{article}
\usepackage[LGR,T1]{fontenc}
\usepackage[utf8]{inputenc}
\usepackage{cprotect}
\usepackage{float}
\usepackage{mathtools}
\usepackage{amsmath}
\usepackage{amssymb}
\usepackage{graphicx}
\usepackage{geometry}
\makeatletter

\DeclareRobustCommand{\greektext}{%
  \fontencoding{LGR}\selectfont\def\encodingdefault{LGR}}
\DeclareRobustCommand{\textgreek}[1]{\leavevmode{\greektext #1}}

\providecommand{\tabularnewline}{\\}

\usepackage{geometry}
\usepackage[sf, bf]{titlesec}

\usepackage{caption}
\makeatletter
\g@addto@macro\bfseries{\boldmath}
\makeatother

\newcommand{\relu}{\mathop{\mathrm{relu}}}
\newcommand{\softmax}{\mathop{\mathrm{softmax}}}
\newcommand{\argmin}{\mathop{\mathrm{argmin}}}
\newcommand{\argmax}{\mathop{\mathrm{argmax}}}
\DeclareMathOperator{\sech}{sech}
\DeclareMathOperator{\sign}{sign}

\usepackage{amsmath}
\let\[\relax \let\]\relax
\DeclareRobustCommand{\[}{\begin{equation}}
\DeclareRobustCommand{\]}{\end{equation}}

\RequirePackage{xcolor}
\definecolor{darkgreen}{rgb}{0, 0.4, 0}
\definecolor{midgreen}{rgb}{0.5, 0.8, 0.5}
\definecolor{darkred}{rgb}{0.6, 0, 0}
\definecolor{darkblue}{rgb}{0, 0, 0.6}
\definecolor{darkgray}{rgb}{0.2, 0.2, 0.2}

\usepackage{tikz}
\usepackage{circuitikz}
\usepackage{tikzscale}  
\usepackage{pgfplots}
\pgfplotsset{compat=1.18}
\usetikzlibrary{backgrounds}
\usepgfplotslibrary{fillbetween}

\makeatletter
\newcount\ctikz@ampsym@n
\def\ctikz@ampsym#1{%
  \global\advance\ctikz@ampsym@n by 1\relax
  \ifnum\ctikz@ampsym@n>2 \else #1\fi}
\def\pgf@circ@ampli@plus{\ctikz@ampsym{\ctikzvalof{amplifiers/plus}}}
\def\pgf@circ@ampli@minus{\ctikz@ampsym{\ctikzvalof{amplifiers/minus}}}
\tikzset{every node/.append style={/utils/exec={\global\ctikz@ampsym@n=0\relax}}}
\makeatother

\RequirePackage{hyperref}
\hypersetup{linktocpage=true, colorlinks=true, citecolor={darkgray}, linkcolor={darkgray}, urlcolor={darkgray} }

\makeatother

\usepackage{listings}

\begin{document}
\title{\bfseries\sffamily\LARGE An all-transistor physical-learning transformer\\
can be trained to perform in-context learning }
\author{\ \\ Michael C. Abbott and Benjamin B. Machta  \\ \normalsize\itshape
Yale Physics \& QBio}
\date{8 October 2026}
\maketitle
\begin{abstract}
\noindent Most systems trained by equilibrium propagation or coupled
learning are in the same class as multi-layer perceptrons: while they
may be universal function approximators, they struggle to learn associative
tasks which are easy for transformers. Equilibrium propagation relies
on a bidirectional flow of information through the physical system
between input and output. The reverse flow is broken by more complex
nonlinear systems, such as those performing multiplication which is
needed to express dot-product attention. Here we explicitly restore
this bidirectional flow by introducing extra circuit elements for
this purpose into a double or dyadic circuit. We then use these ideas
to construct an attention mechanism exploiting the exponential current
law of sub-threshold MOSFET devices. We show that in simulations this
circuit can be trained to perform a simple in-context learning task,
despite using realistic reverse components which transmit only an
approximate gradient. For this training to succeed we need cross-entropy
loss (not mean squared error). We generate nudges appropriate for
this loss using a novel floating circuit element, based on the same
MOSFET sub-circuit used for softmax. We train the resistors in this
circuit using the Lion optimiser, which can be implemented using one
capacitor per parameter to store local memory of past gradient estimates.

\end{abstract}

\section{Introduction}\label{sec:Introduction}

Physical learning is a machine-learning paradigm in which an analogue
physical system performs both the forward computation (say from input
to classification) and also the back-propagation of an error signal
to allow training. In the simplest \emph{equilibrium propagation}
case \cite{scellier2017equilibrium}, an electrical circuit supplied
with some input voltages settles to a \emph{free} steady state, and
some of the resulting node voltages are the output. During training,
the error signal is current injected into the output nodes, causing
the system to find a new \emph{nudged} steady state. At each resistor,
the contrast between free and nudged currents allows us to precisely
reconstruct the gradient of the loss function with respect to this
one conductance, and hence to adjust the resistor so as to perform
gradient descent. This scheme works because the circuit’s equilibrium
states minimise an energy function, namely the power dissipated, subject
to boundary conditions at the input and, during training, at the output. Reducing
the difference in energy between these two states performs gradient
descent.

This simple setup works for systems of reciprocal interactions, including
for instance networks of resistors and diodes. Such networks can easily
be constructed to mimic a multi-layer perceptron (MLP) with relu nonlinearities,
and thus, like MLPs, they are universal function approximators \cite{scellier2024universal}.
However, the ability to \emph{represent} any function does not imply
the ability to \emph{learn} any task. This is why artificial neural
networks moved beyond MLPs and related architectures, and are now
dominated by transformer architectures \cite{vaswani2017attention}.
In a transformer, different parts of the input are compared, most
naturally implemented with transistors, not resistors. In the simplest
cases different inputs, voltages, are eventually multiplied by each
other to perform dot-product attention. By contrast, an MLP only multiplies
inputs by weights, which is what Ohm's law can implement. An ideal
voltage multiplier is a perfect bilinear voltage-controlled current
source (VCCS), which is perfectly non-reciprocal, in the sense that
it completely blocks error signal from propagating back to earlier
circuit elements.

Here, we first explore how to construct modified circuits which allow
the physical back-propagation of error signal through arbitrary nonlinear
elements. We do this with a double or dyadic circuit based on \cite{scurria2026equilibrium,hoier2024dyadic},
in which the steady state is in general a saddle point of some dyadic
potential, rather than a minimum of energy. Instead of using one circuit
for the free case and one for the nudged \cite{dillavou2024machine},
here the mean of each pair of node voltages is the forward computation,
while the difference is the error signal. Extra circuit elements are
added to explicitly pass the error signal backwards, when necessary.
There is an analogy to how automatic differentiation works on the
computer, in which the forward computation is a chain of functions,
and the pullback of each function is the appropriate backward operation
for the error signal.

And second, we show explicitly one possible architecture with transformer-like
behaviour. Rather than precisely implementing the familiar dot-product
attention mechanism, we construct here a version which exploits the
exponential nonlinearity of MOSFET devices in a natural way, replacing
the dot product with a bump circuit \cite{delbruck1991bump} analogous
to Bahdanau attention \cite{bahdanau2016neural}. We implement softmax
by sharing a fixed tail current among outputs controlled by input
gate voltages \cite{elfadel1993softmax}. While the exact dyadic double
of an individual MOSFET is complicated, we show that quite crude approximations
to these larger circuit blocks (using other realistic components)
still allow the double circuit to learn. The learned parameters are
all adjustable resistors, as in the earlier MLP circuits.

Along the way, we show that some other ideas from digital neural networks
can be copied to this analogue world. We implement loss functions
as floating circuit elements connected to the doubled circuit. Mean
squared error is simple, but cross-entropy can also be implemented
using the same MOSFET sub-circuits. And we show how to implement local
optimisation rules smarter than gradient descent. Adam \cite{kingma2017adam}
and Lion \cite{chen2023symbolic} store some memory of past updates,
always for just one parameter, hence they obey the locality of physical
learning.

Finally, to test this model we use a simple in-context learning (ICL)
task, whose input is a string of characters \cite{chan2022datadistributional,reddy2024mechanistic}.
The task, learned from examples, is to take the last character in
the input and compare it to all earlier ones, retrieve part of the
input near to that match, and return this as the output. This is fundamentally
what the learned induction heads in a transformer do \cite{olsson2022context}.
The input characters are encoded as embedding vectors, columns of
an input matrix. The input matrices for this task are smaller than
those of the classic MNIST learning task, of classifying handwritten
digits from 28$\times$28 pixel images \cite{lecun1998mnist}. Yet
this ICL task is very difficult for an MLP to learn, so it serves
as a test that the model truly has new capabilities.

Physical learning is one aspect of a broader revival of interest in
analogue computing, aiming at many-variable machine-learning tasks.
We briefly comment on the relation to other approaches in the conclusion.
And we introduce a conjecture that no physical learning system with
pairwise reciprocal interactions (i.e. one where the state minimises
an energy function) can efficiently learn ICL tasks.

\section{Circuits adapted for physical learning}\label{sec:Circuits-adapted-for-learning}

A multi-layer perceptron (MLP) is a parameterised function $f:\mathbb{R}^{n}\to\mathbb{R}^{m}$,
and with one hidden layer (of dimension $n_{\mathrm{hidden}}$) its
output is
\begin{equation}
y_{c}=\sum^{n_{\mathrm{hidden}}}_{\mu=1}W^{\mathrm{decode}}_{c\mu}\relu\big(\sum^{n}_{a=1}W^{\mathrm{encode}}_{\mu a}x_{a}+b_{\mu}\big)\label{eq:MLP}
\end{equation}
with two matrices $W$ and a bias vector $b$ of adjustable parameters.
Here $\relu(z)=\max(z,0)$ is a common choice of activation function,
although many others have been used. If we set $n=m=1$, then it is
not surprising that any smooth function $g:[-1,1]\to[-1,1]$ can be
approximated by this formula. The approximation $f$ is piecewise
linear, and each hidden node $\mu=1,2,\ldots n_{\mathrm{hidden}}$
adds another kink whose position and slope can be adjusted, so that
for large $n_{\mathrm{hidden}}$ we can make $\int dx[f(x)-g(x)]^{2}$
small. This can be proven for any dimensionality and a wide class
of activation functions \cite{cybenko1989approximation,leshno1993multilayer}.

\begin{figure}
\centering \mbox{\includegraphics[width=10cm]{figures/dense3.tikz}\hspace{0.5cm}\includegraphics[height=4.5cm]{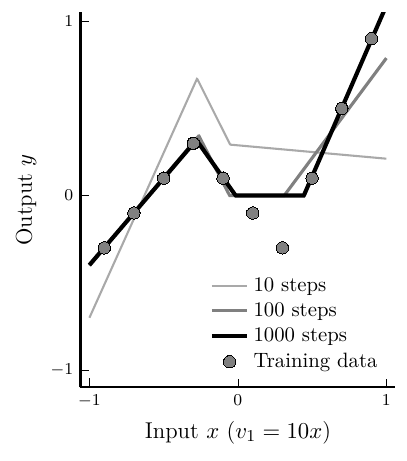}}

\caption{Circuit simulating a relu-MLP, similar to \cite{kendall2020training,scellier2024universal},
with $n=1$ inputs supplied as $v_{1,2}=\pm Ax_{1}$ and one output
$y=v_{8}$. The hidden nodes have alternating diodes to/from ground,
enforcing $v_{5}\protect\geq0$, $v_{6}\protect\leq0$, $\ldots$,
and the input $x$ at which the diodes turn on or off can be adjusted
using the resistors to $v_{3,4}=\pm B$, which are fixed bias nodes.
Closing the switch introduces nudge current for training, calculated
from (\ref{eq:nudge-single}). Plot at right shows steps of training
this circuit by gradient descent to match the datapoints shown, by
adjusting all resistors. The flat segment $y=0$ where all diodes
are off is a simple example of a circuit failing to learn a function
it is able to express. }\label{fig:Single-MLP}
\end{figure}

Figure \ref{fig:Single-MLP} shows a circuit implementation of this
MLP, in which adjustable resistors encode the parameters. The input
is supplied by fixing node voltages $v_{1}$ to $v_{n}$, but because
resistance cannot be negative, we supply also inverted inputs. The
diodes connecting the hidden nodes to ground implement relu nonlinearity,
but we similarly need them in both directions to implement $\max(v_{b},0)$
and $\min(v_{b},0)$ separately. More subtly, while entries in $W^{\mathrm{decode}}$
can be large allowing any linear combination, the output resistors
cannot amplify. For this reason the input nodes are amplified by a
factor $A>1$, to fix $2n$ voltages as follows:
\[
v_{a}=Ax_{a},\qquad v_{a+n}=-Ax_{a},\qquad a=1,2,\ldots n.
\]
The output is read directly from some other voltages:
\[
y_{c}=v_{c},\qquad c=N,N-1,\ldots,N-m+1.
\]
For $n=m=1$, this circuit gives a piecewise linear $\mathbb{R}\to\mathbb{R}$
function $y=f(x)$. At each kink a diode turns on or off. (But the
number of kinks can be larger than the number of diodes, as the hidden
nodes are shifted by the output layer --- see figure \ref{fig:Five-kinks}
in the appendix.) The function obeys a slope constraint $\left|f'(x)\right|\leq A$,
and the universal approximation results of \cite{scellier2024universal}
for this circuit take both the amplification factor $A$ and the number
of nodes $N$ to be large.

This circuit can be trained using the equilibrium propagation scheme
of Scellier and Bengio \cite{scellier2017equilibrium}. First the
input is supplied, and we write $v^{\mathrm{free}}_{a}$ for the resulting
steady-state node voltages. If the desired output is $y^{\mathrm{target}}$
and the loss function is $L(y,y^{\mathrm{target}})$, then the nudged
state is made by injecting the following current into the output nodes
(closing the switch in figure \ref{fig:Single-MLP}):
\begin{equation}
i^{\mathrm{nudge}}_{0\to c}=-\beta\:\frac{\partial L(y,y^{\mathrm{target}})}{\partial y_{c}}.\label{eq:nudge-single}
\end{equation}
Here $\beta$ sets the scale of the nudge, and $0\to c$ indicates
that the current flows from ground. Consider a resistor between nodes
$a$ and $b$ with conductance $g_{ab}=1/r_{ab}$. If its voltage
drop is $\Delta v_{ab}=v_{a}-v_{b}$, then gradient of the loss with
respect to its conductance can be estimated by comparing the free
and nudged voltage drops \cite{scellier2017equilibrium}:
\begin{equation}
\frac{dL}{dg_{ab}}\approx\frac{1}{2\beta}\left[(\Delta v^{\mathrm{nudged}})^{2}-(\Delta v^{\mathrm{free}})^{2}\right].\label{eq:gradient-nudge}
\end{equation}
Gradient descent with learning rate $\eta>0$ is then, as usual, this
update:
\begin{equation}
g_{ab}\mapsto g_{ab}-\eta\:\frac{dL(y,y^{\mathrm{target}})}{dg_{ab}}.\label{eq:gradient-descent}
\end{equation}
Such an update can be performed locally, seeing only the voltages
at this particular resistor (provided $\Delta v^{\mathrm{free}}$
can be memorised until the nudged phase). This local learning property
is the key attraction of equilibrium propagation and other physical
learning schemes. All of the transport of information, in both directions,
happens in the circuit.

A variant of this scheme introduced by Stern et al. \cite{stern2021supervised}
clamps the output nodes to fixed voltages, instead of injecting a
fixed nudge current. If we observe voltage $y^{\mathrm{free}}_{c}$
in the free state, then the clamped state imposes
\begin{equation}
v_{c}=(1-\gamma)y^{\mathrm{free}}_{c}+\gamma y^{\mathrm{target}}_{c},\qquad\gamma\ll1.\label{eq:nudge-clamp}
\end{equation}
The voltage source used to fix this output node injects a current
proportional to $\gamma(y^{\mathrm{target}}_{c}-y^{\mathrm{free}}_{c})$,
so at each node $c$ this amounts to choosing $\beta$ scaled by the
output impedance: $\beta_{c}\approx\gamma\:\nu^{2}/R^{\mathrm{out}}_{c}$.
(Here $\nu$ is the voltage scale of the loss, needed for units, see
below.)

Related two-phase training schemes have a long history, originally
applied to Boltzmann machines \cite{ackley1985learning} and Hopfield
networks \cite{movellan1991contrastive} under the name contrastive
Hebbian learning, and later to MLPs \cite{oreilly1996biologically,xie2003equivalence}.

\subsection{Double and dyadic circuits}\label{subsec:Double-and-dyadic}

Instead of allowing the circuit to settle to free and nudged states
at different times, we now introduce a double circuit which does both
at once. But unlike the twin circuits of Dillavou et al. \cite{dillavou2024machine},
where free and nudged run on different copies of the hardware, in
this double (or dyadic) circuit the free state and the nudge are the
mean and difference of the two copies, as in \cite{hoier2023dual,scurria2026equilibrium}.

\begin{figure}
\centering \includegraphics[width=13cm]{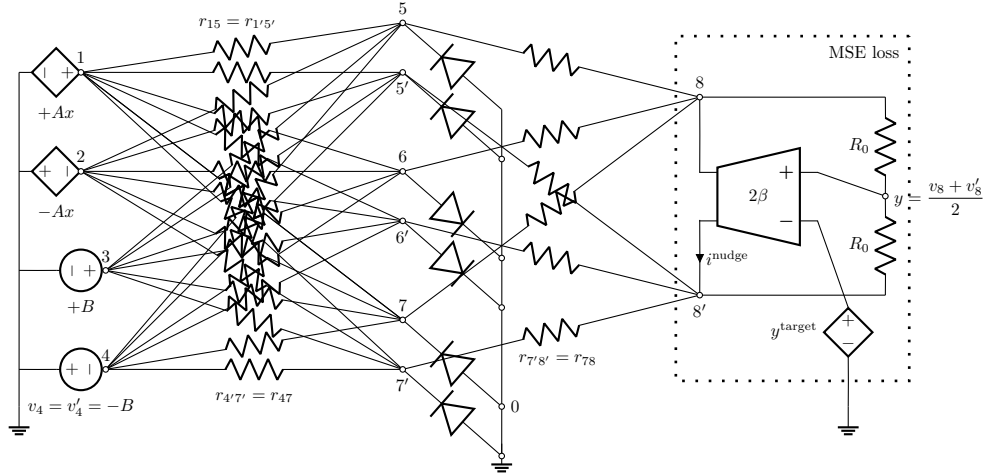}

\caption{Double version of the circuit in figure \ref{fig:Single-MLP}, equipped
with a floating loss function. Every node and every resistor or diode
is duplicated, but the input clamps for instance $v_{1}=v_{1}'=Ax$
so we identify $1=1'$. The output is now $y=(v_{8}+v_{8}')/2$, explicitly
computed by a voltage divider with large fixed $R_{0}$. For training,
the gradient of MSE loss is provided by a VCCS (fully differential
OTA), which compares this output voltage to $y^{\mathrm{target}}$
to create the nudge current $i_{8\to8'}\propto\beta$, and ultimately
produces $v_{a}-v_{a}'\propto\beta$ at other nodes. The circuit has
the same 15 parameters as before, as we maintain $r_{ab}=r_{a'b'}$
during training. An exactly dyadic circuit could not use ordinary
diodes, and would instead couple their behaviour so that $i_{5\to0}$
and $i_{5'\to0}$ are either both zero (off) or both nonzero (on).
}\label{fig:Double-MLP}
\end{figure}

Figure \ref{fig:Double-MLP} shows a doubled version of the MLP-like
circuit in figure \ref{fig:Single-MLP}. The new circuit has $2N$
nodes, with voltages $v_{a}$ and $v_{a}'$. The input fixes $4n$
of them:
\[
v_{a}=v_{a}'=Ax_{a},\qquad v_{a+n}=v_{a+n}'=-Ax_{a},\qquad a=1,2,\ldots n
\]
and the $m$ outputs are constructed using $2m$ voltages, by taking
the mean of original and primed copies:
\[
y_{c}=\frac{v_{c}+v'_{c}}{2},\qquad c=N,N-1,\ldots.
\]
Every resistor exists in two copies $g_{ab}=g_{a'b'}$, and every
diode, with no cross-connections.

What connects the two copies is that the training nudge is now delivered
between original and primed output nodes:
\begin{equation}
i^{\mathrm{nudge}}_{c\to c'}=\beta\:\frac{\partial L(y,y^{\mathrm{target}})}{\partial y_{c}}\label{eq:nudge-double}
\end{equation}
producing differences $v_{a}-v_{a}'\propto\beta$ elsewhere. The gradient
is read from the difference in how the two copies behave:
\begin{equation}
\frac{dL}{dg_{ab}}\approx\frac{1}{4\beta}\left[(\Delta v_{ab})^{2}-(\Delta v_{a'b'})^{2}\right].\label{eq:gradient-double}
\end{equation}
For a circuit of resistors and diodes (such as the one in figure \ref{fig:Double-MLP}),
this gradient formula is in fact exact for sufficiently small $\beta$,
small enough that no diode turns on or off.

So far this circuit of reciprocal components might be viewed as a
circuit version of Høier et al. \cite{hoier2023dual}, which was expanded
in \cite{hoier2024dyadic} to allow linear non-reciprocal interactions.
This was generalised to arbitrary differentiable dynamical systems
by Scurria et al. \cite{scurria2026equilibrium}, and we will describe
as ``dyadic'' those circuits constructed according to their rules,
while using ``double'' for a wider class of circuits with paired
nodes obeying (\ref{eq:gradient-double}). The dyadic rules make the
gradient estimate (\ref{eq:gradient-double}) true for any $\beta$,
thus the circuit in figure \ref{fig:Double-MLP} is not in fact exactly
dyadic (in this usage). This is because its diodes switch independently,
while the dyadic counterpart of a single diode is two diodes locked
to be both on, or both off, according to the mean of their voltage
differences --- see appendix \ref{subsec:Diode}.

This circuit is also an implementation of the symmetric nudging idea
of Laborieux et al. \cite{laborieux2020scaling}, who showed that
two nudge phases of strengths $\pm\beta$ resulted in more precise
gradient estimation. Before returning to questions of non-reciprocal
elements, we show how the nudge may be physically generated within
the circuit's analogue world, so that there is only one phase of training,
not two or three.

\subsection{Loss as a circuit component}\label{subsec:Loss}

Suppose the loss function is mean squared error (MSE):\footnote{Note aside that the clamped training scheme (\ref{eq:nudge-clamp}),
voltage mixing with $\gamma\ll1$, is equivalent to a nudge of strength
$\beta=\gamma\:\nu^{2}/\rho$ following from this loss function:
\begin{equation}
L_{\mathrm{clamp}}(y,y^{\mathrm{target}})=\frac{\rho}{2\nu^{2}}\sum^{m}_{c,d=1}(y_{c}-y^{\mathrm{target}}_{c})G_{cd}(y_{d}-y^{\mathrm{target}}_{d})\label{eq:loss-clamp}
\end{equation}
where $G$ is the Thevenin conductance matrix at the output nodes,
evaluated at the free operating point, and $\nu,\rho$ are voltage
and resistance references to make loss dimensionless. This equivalence
should be valid for any reciprocal circuit, not just resistors and
diodes, but may be broken by the active circuits below.}
\begin{equation}
L_{\mathrm{MSE}}(y,y^{\mathrm{target}})=\frac{1}{m\nu^{2}}\sum^{m}_{c=1}(y_{c}-y^{\mathrm{target}}_{c})^{2}\label{eq:loss-MSE}
\end{equation}
where $\nu=1$V ensures the loss is dimensionless. Then the nudge
current for the doubled circuit is this:
\begin{equation}
i^{\mathrm{nudge}}_{c\to c'}=\frac{2\beta}{m\nu^{2}}\Big(\frac{v_{c}+v'_{c}}{2}-y^{\mathrm{target}}_{c}\Big).\label{eq:nudge-MSE}
\end{equation}
If we regard $y^{\mathrm{target}}_{c}$ as the voltage of an additional
node, then the nudge is one more circuit element: a linear voltage-controlled
current source (VCCS), i.e. a fully differential operational transconductance
amplifier (OTA). This is drawn in figure \ref{fig:Double-MLP}. With
such a circuit element inserted, the training scheme now has only
one phase: the input and target voltages are applied simultaneously,
and when the double circuit settles to a steady state, the gradient
can be immediately estimated from (\ref{eq:gradient-double}). We
do not need to read the free behaviour, and then decide what nudge
to apply, from outside.

Other loss functions can similarly be implemented as floating circuit
elements. Cross-entropy loss is the usual choice for classification
tasks such as MNIST, and was used by \cite{laborieux2020scaling,kerjan2026training}
for equilibrium propagation (training circuits equivalent to convolutional
neural networks). For this loss function, output $y\in\mathbb{R}^{m}$
from a neural network is first transformed to probabilities $p^{\mathrm{out}}(y)\in\mathbb{R}^{m}$
by softmax. Our version includes a temperature $\tau$:
\begin{equation}
p^{\mathrm{out}}(y)_{c}=\softmax(y/\tau)_{c}=\frac{e^{y_{c}/\tau}}{\sum_{d}e^{y_{d}/\tau}}\label{eq:softmax}
\end{equation}
and clearly $\sum_{c}p^{\mathrm{out}}(y)_{c}=1$. This distribution
is then compared to a supplied target distribution $p^{\mathrm{target}}$
using cross-entropy, giving the following loss function:
\begin{equation}
L_{\mathrm{LCE}}(y,p^{\mathrm{target}})=-\sum_{c}p^{\mathrm{target}}_{c}\log p^{\mathrm{out}}(y)_{c}.\label{eq:loss-LCE}
\end{equation}
The raw outputs $y$ are termed logits, and the function $L_{\mathrm{LCE}}$
which accepts them is logit cross-entropy (LCE). This loss function
penalises low probabilities of the correct label $c^{\mathrm{target}}$
more heavily than high probabilities of incorrect labels. And it is
invariant to overall shifts $y_{c}\mapsto y_{c}+\mathrm{const}$.
Such shifts do not affect the identity of the largest output, $\text{argmax}_{c}y_{c}$,
hence do not affect classification accuracy.

The nudge current required for LCE is $i^{\mathrm{nudge}}_{c\to c'}=(\beta/\tau)\left[\softmax(y/\tau)_{c}-p^{\mathrm{target}}_{c}\right].$
We have an implementation of this as a floating circuit element, exploiting
the exponential current law of the MOSFETs to implement softmax using
same ``bank'' circuit shown in figure \ref{fig:Bank}. We describe
the full circuit in appendix \ref{sec:Cross-entropy-loss}.

\subsection{Perfect voltage multiplier}\label{subsec:Perfect-voltage-multiplier}

Now we wish to introduce nonlinear, non-reciprocal circuit elements,
while still allowing for the bidirectional flow of information. Rather
than thinking about slight perturbations around the above case, we
proceed immediately to the opposite extreme, by inserting another
ideal voltage-controlled current source.

Voltages can be multiplied by a circuit of 6 transistors called a
Gilbert cell \cite{gilbert1968precision}, which measures two pairs
of voltage differences $v_{a}-v_{b}$ and $v_{c}-v_{d}$, and then
drives current between two additional nodes, say $v_{e}$ and $v_{f}$.
The simplest ideal version is a voltage-controlled current source,
with current law
\begin{equation}
i_{e\to f}=K\:\Delta v_{ab}\:\Delta v_{cd}\label{eq:current-Gilbert}
\end{equation}
where $K$ is a fixed gain. Besides these 6 terminals, the cell must
of course be connected to a power supply.

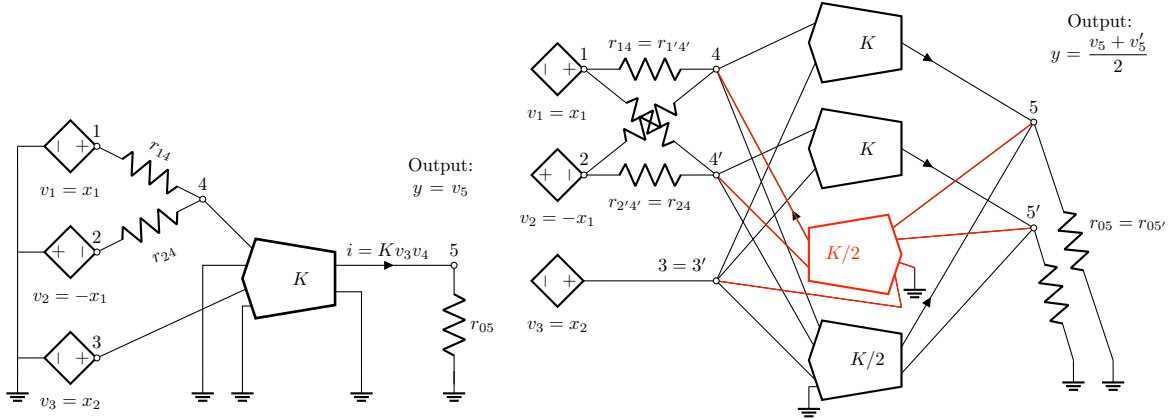
\begin{figure}
\centering \input{figures/denver.tikz}\hspace{1mm}\input{figures/denver_double_red.tikz}

\caption{Simple multiplier circuit using an ideal Gilbert cell, a six-terminal
device with three connected to ground here. Nudging the output $y=v_{5}$
of the circuit on the left cannot perturb node $v_{4}$, before the
multiplier. In the dyadic version of the same circuit (on the right),
each node and each resistor is duplicated, but the multiplier is replaced
with four multipliers: one in each copy, one reversed, and one providing
a correction term. (Some connections to ground are not shown.) With
this configuration, a nudge of order $\beta$ applied to $i_{5\to5'}$
will preserve all mean node voltages $(v_{a}+v_{a}')/2$ exactly,
while $v_{4}-v'_{4}\propto\beta$ has gradient information about $dL/dr_{14}$
and $dL/dr_{24}$.}\label{fig:Denver}
\end{figure}

A simple circuit containing one Gilbert cell and three resistors is
drawn in figure \ref{fig:Denver}, whose output voltage is $y=Kr_{50}\frac{r_{24}-r_{14}}{r_{14}+r_{24}}x_{1}x_{2}$.
Here the Gilbert cell has half its nodes connected to ground, so its
current law is
\begin{equation}
i_{0\to5}=K\:v_{3}\:v_{4}.\label{eq:current-Gilbert0}
\end{equation}
It should be obvious that a nudge of the output node in figure \ref{fig:Denver}
cannot have any effect on the part of the circuit left of the Gilbert
cell. Hence $\Delta v^{\mathrm{free}}_{14}=\Delta v^{\mathrm{nudged}}_{14}$,
so we gain no information about $dL/dg_{14}$, and cannot tune resistor
$r_{14}$.

This problem can be fixed in the doubled circuit by inserting (in
addition to the obvious duplicate) additional multipliers, some facing
the other way. To derive what is required, we construct the dyadic
potential of \cite{scurria2026equilibrium}. At each free node $a$,
this multiplies $(v_{a}-v'_{a})$ by the current leaving the node.
For the circuit in figure \ref{fig:Denver} it reads
\[
H(v,v')=\frac{g_{14}}{2}(v^{2}_{4}-v^{\prime2}_{4})+\frac{g_{24}}{2}(v^{2}_{4}-v^{\prime2}_{4})\underbrace{-(v_{5}-v_{5}')\frac{K}{4}(v_{3}+v_{3}')(v_{4}+v_{4}')}_{=h(v,v')}+\frac{g_{50}}{2}(v^{2}_{5}-v^{\prime2}_{5})
\]
Each term arises from a component, and the currents required in the
dyadic circuit may be read off as derivatives of this potential. For
the Gilbert cell, these currents are:
\[
\begin{aligned}i_{0\to5}=-\frac{\partial h}{\partial v_{5}} & =\frac{K}{4}(v_{3}+v_{3}')(v_{4}+v'_{4}) & i_{0\to5'}=+\frac{\partial h}{\partial v'_{5}} & =\frac{K}{4}(v_{3}+v_{3}')(v_{4}+v'_{4}) &  & \text{forward}\\
i_{0\to4} & =\frac{K}{4}(v_{3}+v_{3}')(v_{5}-v'_{5}) & i_{0\to4'} & =-\frac{K}{4}(v_{3}+v_{3}')(v_{5}-v'_{5}) &  & \text{backward}.
\end{aligned}
\]
There would be similar currents into nodes $3,3'$ if these were not
fixed. Assuming $v_{3}=v'_{3}$, the two backward current terms are
simple to implement, as a single 6-terminal Gilbert cell can sense
$v_{5}-v'_{5}$ and drive $i_{4\to4'}$, to give
\begin{equation}
i_{4'\to4}=(K/2)\:v_{3}\:\Delta v_{55'}.\label{eq:gilbert-adjoint1}
\end{equation}
For the forward currents, expanding $v_{3}(v_{4}+v'_{4})=v_{3}v_{4}+v_{3}v_{4}'$
gives a set of four forward multipliers, each working relative to
ground. But the same currents can be produced by just three multipliers
--- the original one, its obvious double, and a correction term:
\begin{align}
i_{0\to5} & =K\:v_{3}\:v_{4}\nonumber \\
i_{0\to5'} & =K\:v_{3}\:v_{4}'\label{eq:gilbert-adjoint3}\\
i_{5\to5'} & =(K/2)\:v_{3}\:\Delta v_{44'}\nonumber
\end{align}
These three forward multipliers, and one reverse (\ref{eq:gilbert-adjoint1}),
are drawn in figure \ref{fig:Denver}.

We review this dyadic construction at greater length in appendix \ref{sec:Deriving-current-rules},
and introduce an approximation which drops terms $\mathcal{O}(\beta^{2})$
(but is exact for case $v_{3}=v'_{3}$ above). It is always possible
to specify a dyadic circuit using the rules of \cite{scurria2026equilibrium},
but there is no guarantee that it can be implemented from the same
components --- that is a special property of multiplication. Arguably
it's the same special property which makes resistors their own dyadic
doubles, with no extra backward components. The next section explores
this in an informal way.

\subsection{Aside on automatic differentiation}\label{subsec:Aside-on-AD}

The reverse multiplier introduced in figure \ref{fig:Denver} is reminiscent
of the reverse operations inserted computing gradients in digital
neural networks --- we might guess the connections, if not the $K/2$
factor. This dyadic construction is deeply connected to automatic
differentiation, but not quite so simply, because the different ``layers''
of the circuit from input to output aren't just functions acting one
after the other. This section reviews reverse-mode automatic differentiation
in its simplest form, to gain intuition.

Given a function $y=f(x)$ and a scalar loss function $L(y)$, the
essence of back-propagation is to transform $\bar{y}=dL/dy$ into
$\bar{x}=dL/dx$.\footnote{The notation is standard, but be warned that these gradients are $v-v'$
in the dyadic circuits, not the mean $(v+v')/2$ as the bar might
suggest. } If we regard $x$ and $y$ as living on two different manifolds,
then the map required is called the \emph{pullback} of $f$ in differential
geometry, written $\bar{x}=f^{*}(\bar{y})$. This is always linear
(it's a homomorphism between cotangent vector spaces, $T^{*}_{y}\to T^{*}_{x}$,
\cite{cartan1923lecons,whitney1957geometric}) but it may depend on
$x$. One simple example is:
\[
y=f(x)=\tanh x\qquad\Longrightarrow\qquad\bar{x}=f^{*}(\bar{y})=[1-(\tanh x)^{2}]\:\bar{y}.
\]
Given a long chain of functions, like $z=h(g(f(x)))$, the gradient
(or error signal) $\bar{z}=dL/dz$ can be pulled back through each
in turn, $\bar{x}=f^{*}(g^{*}(h^{*}(\bar{z})))$.

This procedure was used early on for digital simulations of analogue
circuits \cite{hachtel1969costate}, as well as for neural networks
\cite{amari1967theory,lecun1992theoretical}. Computer scientists
stress is that the complexity of finding the gradient is $\mathcal{O}(1)$
times that of the original function, rather than growing linearly
with the number of inputs \cite{speelpenning1980compiling}. This
is also true of our doubled circuits.

If the function is multiplication, $f(x_{1},x_{2})=x_{1}\:x_{2}$,
then there are two pullbacks, one for each argument, and both involve
multiplication:
\begin{align}
y=f(x_{1},x_{2}) & =x_{1}\:x_{2}\qquad\Longrightarrow & \bar{x}_{1}=f^{*}_{1}(\bar{y}) & =\bar{y}\:x_{2}{}^{\intercal} & \bar{x}_{2}=f^{*}_{2}(\bar{y}) & =x_{1}{}^{\intercal}\:\bar{y}.\label{eq:autodiff-multiply}
\end{align}
That is precisely the form seen above, where $\Delta v_{55'}$ plays
the role of $\bar{y}$: one input of the reverse multiplication is
the error signal of the output, while the other ($v_{3}$ above) is
part of the input.

The circuit law for resistors $i_{ab}=g_{ab}\Delta v_{ab}$ is also
multiplication, and this is why its dyadic circuit element is just
the same resistor again. Something similar is true for diodes, which
are (at any given operating point) a limiting case in which the resistance
is either $0$ or $\infty$. However, the effective resistance of
both diodes should depend on the same $\Delta v$, hence an independent
pair of diodes (as in figure \ref{fig:Double-MLP}) is not quite the
exact dyadic double.

\subsection{Ideal sub-threshold MOSFET}\label{subsec:maintext-MOSFET}

An ideal transistor conducts current between two nodes, source and
drain, with a conductance controlled by voltage at a third, the gate.
There are many types, and they obey relatively simple laws in particular
limits of their node voltages. Within physical learning, MOSFETs in
the triode regime have been used as slightly nonlinear adjustable
resistors, with their gates connected to a capacitor to control the
resistance \cite{dillavou2024machine}. We're instead going to use
them in the sub-threshold (or weak inversion) regime, where current
varies exponentially with the voltages, because these exponentials
will allow us to implement attention below.

Our ideal device is a 3-terminal component (gate, source, drain) with
the following current law \cite{enz1995analytical}:
\begin{equation}
i_{d\to s}(v_{g},v_{s},v_{d})=I_{0}e^{(v_{g}-V_{\mathrm{th}})/nV_{T}}\left[e^{-v_{s}/V_{T}}-e^{-v_{d}/V_{T}}\right]\label{eq:current-mosfet}
\end{equation}
with constants $V_{T}=0.026$V, $V_{\mathrm{th}}=0.7$V, $I_{0}=10^{-8}$A,
$n=1.5$. No current flows through the gate. This law is valid for
voltages $v_{g}<V_{\mathrm{th}}+n\,\min(v_{s},v_{d})$ , or equivalently,
current $\left|i_{d\to s}\right|<I_{0}$.

Below we use three different circuits in which a current source is
connected to the source terminal of several transistors, shown in
figures \ref{fig:Bump}, \ref{fig:Bank}, and \ref{fig:Pair}. Their
gates control how this current is shared among their drains, and the
exponentials give us useful functional forms.

The dyadic circuit component necessary for a MOSFET to pass error
signal backwards is quite complicated. What we do below is, instead,
to construct more realistic approximately dyadic versions of these
larger blocks. Quite crude approximations, much worse than $\mathcal{O}(\beta^{2})$,
are nevertheless sufficient for learning.

\subsection{Adaptive, local, optimisation rules}\label{subsec:maintext-optimisation}

The scheme above uses gradient descent (\ref{eq:gradient-descent})
to update parameters, but this is the most naiive algorithm for using
the gradient to perform optimisation. Better algorithms often try
to estimate the Hessian of the loss function using some memory of
past gradients \cite{nocedal1980updating}, but for efficiency, most
of the algorithms used with neural networks estimate only the diagonal
of the Hessian. This means that they obey the locality criterion of
physical learning. The most famous member of this class is probably
Adam \cite{kingma2017adam}, and we show in appendix \ref{sec:Adaptive-optimisation-rules}
how to implement the simpler Lion \cite{chen2023symbolic} which stores
just one number per parameter. These adaptive rules are invariant
to the scale of the gradients (they have learning rate $\eta$ with
units of conductance, not conductance-squared) which will be important
in circuits with exponential voltage dependence -- like the transformer
we construct in the next section.

\begin{figure}
\centering \includegraphics[width=10cm]{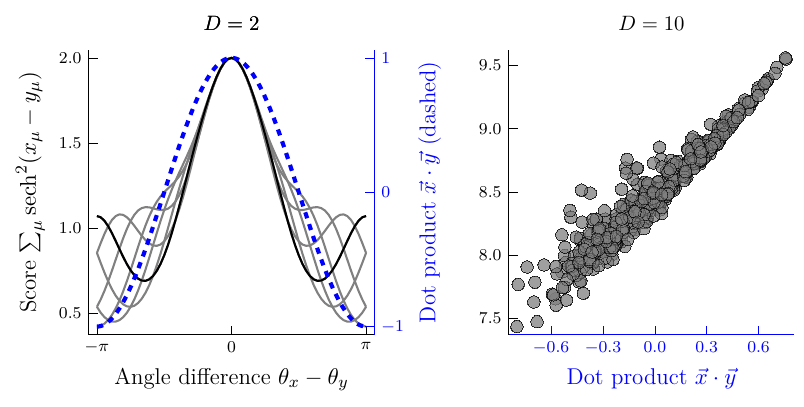}\caption{Comparison of a Bahdanau-like score function \cite{bahdanau2016neural}
to the dot product, applied to unit vectors in two dimensions (left)
and ten dimensions (right). In two dimensions, the vectors are $\vec{x}=(\cos\theta_{x},\sin\theta_{x})$
and $\vec{y}=(\cos\theta_{y},\sin\theta_{y})$, and we sweep $\theta_{x}$
for various fixed $\theta_{y}$, with $\theta_{y}=0$ in bold. In
ten dimensions, we show pairwise comparisons between 30 different
random unit vectors. The claim is that this captures much of what
the dot product captures, although it lacks $SO(D)$ rotational invariance.}\label{fig:Bahdanau}
\end{figure}

\section{Transistor transformer circuit}\label{sec:Transformer-circuit}

Every transformer is a sequence model, which we will write as $f(X)=y$,
where the input $X$ is a matrix whose columns $\vec{x}_{t}$ encode
successive tokens in the input sequence, $t=1,2,\ldots T$, and the
output $y$ gives a probability distribution for the next token. The
core operation is dot product self attention \cite{vaswani2017attention},
which in the simplest (single-headed) case looks like this, with three
weight matrices $W$:
\begin{align}
S_{st} & =(W^{\mathrm{key}}\vec{x}_{s})\cdot(W^{\mathrm{query}}\vec{x}_{t})\quad=\sum_{\mu,\nu,\sigma}X_{\nu s}W^{\mathrm{key}}_{\sigma\nu}W^{\mathrm{query}}_{\sigma\mu}X_{\mu t} &  & \text{scores}\nonumber \\
A_{st} & =\softmax_{s}\begin{cases}
S_{st} & s\leq t\\
-\infty & \text{else}
\end{cases}\quad=\frac{\exp(S_{st})}{{\textstyle \sum}_{s'\leq t}\exp(S_{s't})} &  & \text{attention weights}\label{eq:attention}\\
Y_{\mu t} & =\sum_{\nu}W^{\mathrm{value}}_{\mu\nu}\sum_{s\leq t}X_{\nu s}A_{st} &  & \text{value mixing.}\nonumber
\end{align}
Here we write vector indices $\mu,\nu=1,2,\ldots D$; often the range
of $\sigma$ is smaller making the matrix $(W^{\intercal}_{\mathrm{key}}W_{\mathrm{query}})$
low-rank. The output for the last token is $y_{\mu}=Y_{\mu T}$, and
if this is what we want, then we need only compute elements $A_{sT}$,
with $s=1,2,\ldots T$.

Our goal here is to translate these pieces to active circuit elements,
and then to construct a double circuit with approximate reverse elements,
so that the weights $W$ can be learned. One important approximation
we make is to replace the dot product with something like Bahdanau
attention \cite{bahdanau2016neural}:
\begin{equation}
S_{st}=S_{0}\sum_{\mu}\sech^{2}\left[(W^{\mathrm{key}}\vec{x}_{s})_{\mu}-(W^{\mathrm{query}}\vec{x}_{t})_{\mu}\right].\label{eq:score-Bahdanau}
\end{equation}
This is largest when the key and query vectors $(\ldots)_{\mu}$ agree,
and declines as any components begin to disagree, although it never
becomes negative --- see figure \ref{fig:Bahdanau}. We will also
simplify $W^{\mathrm{key}}\vec{x}_{s}$ by assuming this weight matrix
is diagonal with positive entries, on the grounds that other transformations
can be absorbed into $W^{\mathrm{query}}$.

\begin{figure}
\centering\mbox{\hspace{-2mm}\raisebox{4mm}{\input{figures/bump.tikz}}\includegraphics[width=10cm]{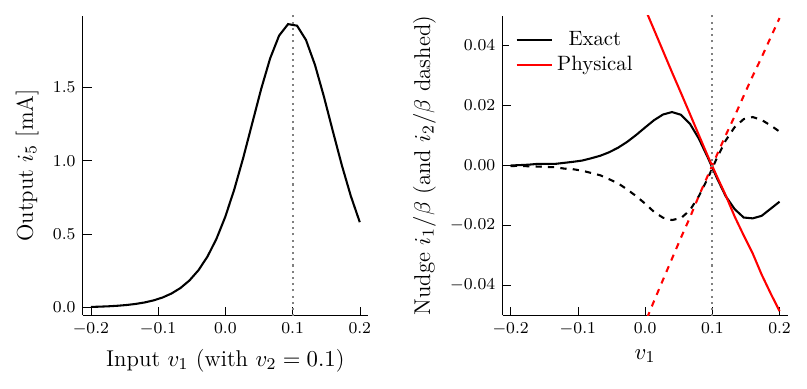}}\caption{Delbrück's bump circuit \cite{delbruck1991bump}, which we use to
replace the dot product. Here $I_{\mathrm{tail}}$ is divided three
ways by four transistors, with the one output current gated by both
inputs. Plots show the output (varying $v_{1}$ at fixed $v_{2}$),
and the effect of nudging output $i_{5}$ on the inputs $i_{1}$,
$i_{2}$. The exact dyadic circuit produces the expected gradient,
while the crude approximation we use with Gilbert cells (in red) differs
greatly away from the centre of the bump, but nevertheless allows
training to move towards or away from the bump as necessary. (These
plots rely on embedding the bump in a larger circuit, and thus depend
on impedance of this apparatus.) }\label{fig:Bump}
\end{figure}

\begin{figure}
\centering \raisebox{2mm}{\input{figures/bank.tikz}}\hspace{1cm}\includegraphics[width=10cm]{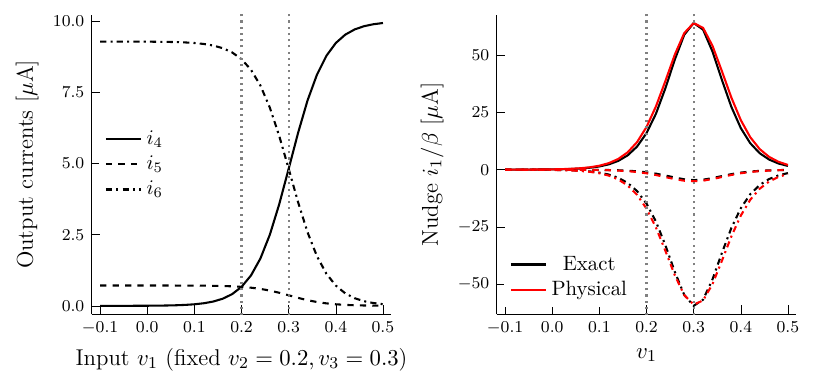}\caption{Three transistors dividing fixed $I_{\mathrm{tail}}$ to produce currents
which behave like softmax of the input voltages \cite{elfadel1993softmax,sillman2023analog},
in a circuit we call ``bank'' . The first plot shows its three outputs
as $v_{1}$ is varied, holding fixed $v_{2},v_{3}$. The second plot
shows one column of the Jacobian, i.e. the error signal measured at
the first input when each of the three outputs is perturbed by a nudge
current. We find it necessary for training to use a fairly elaborate
reverse circuit block which closely follows the ideal dyadic behaviour.
}\label{fig:Bank}
\end{figure}

The learned weights $W$ are implemented by arrays of resistors, much
like the earlier MLP layer in figure \ref{fig:Single-MLP}. The nonlinear
parts use transistor circuits shown in figures \ref{fig:Bump}, \ref{fig:Bank}, and \ref{fig:Pair},
and we now discuss them one by one.

\subsection{Scores $S_{st}$ as currents from ``bump''}

Delbrück’s simple bump circuit \cite{delbruck1991bump} has four transistors
controlled by two gate voltages, $v_{1},v_{2}$, and gives a similarity
score as a current. A fixed current source $I_{\mathrm{tail}}$ is
split into three paths: one gated by $v_{1}$, one gated by $v_{2}$,
and a third limited by two transistors in series, whose gates are
$v_{1}$ and $v_{2}$, which is the output --- node 5 in figure \ref{fig:Bump}.
When the voltages $v_{1}$ and $v_{2}$ are equal, current flows through
all three paths, with 20\% to the output. But when either voltage
is increased or decreased, one of the drain paths is favoured, decreasing
the output. After some algebra, the output current is:
\begin{equation}
i_{5\to0}=\frac{I_{\mathrm{tail}}}{1+4\cosh^{2}\left[(v_{1}-v_{2})/(2nV_{T})\right]}.\label{eq:current-bump1}
\end{equation}
With two vectors of voltages, we can use $D$ such circuits, and add
the resulting currents:
\[
i_{\mathrm{bump}}(\vec{v},\vec{\tilde{v}})=\sum_{\mu}\frac{I_{\mathrm{tail}}}{1+4\cosh^{2}\left[(v_{\mu}-\tilde{v}_{\mu})/(2nV_{T})\right]}
\]
This will play the role of $S_{st}$ above, with $\vec{v}=W^{\mathrm{query}}\vec{x}_{t}+V_{\mathrm{bias}}$
implemented as a crossbar array with inverted inputs: $2D^{2}$ resistors
(as used for the MLP above). We take $W^{\mathrm{key}}$ to be diagonal
and all-positive, thus $\tilde{v}_{\mu}=W^{\mathrm{key}}_{\mu\mu}X_{\mu s}+V_{\mathrm{bias}}$,
implemented as a resistor divider to ground for each $\mu$, thus
$2D$ resistors for each $s$. We need to apply the same array of
resistances to every $\vec{x}_{s}$, which we do with $2DT$ resistors
in which each group of $T$ must be updated together during training.
Finally, between these resistor arrays and the bump circuit, we add
a fixed voltage $V_{\mathrm{bias}}$, which cancels out of (\ref{eq:current-bump1})
but increases the current through each transistor.

In the doubled circuit, there is a second copy of the bump circuit,
with $i_{0\to5'}(v_{1}',v_{2}')$ obeying the same law. We still need
to pass gradient signal from $v_{5}-v_{5}'$ back to the inputs, to
ultimately train $W^{\mathrm{query}}$ and $W^{\mathrm{key}}$. The
sign of this signal must change as we pass from $v_{1}<v_{2}$ to
$v_{1}>v_{2}$. One simple idea is to use a Gilbert cell, or rather
four of them:
\[
i_{1'\to1}=i_{2\to2'}=K_{\mathrm{bump}}(v_{1}-v_{2})(v_{5}-v_{5}')+K_{\mathrm{bump}}(v_{1}'-v_{2}')(v_{5}-v_{5}')
\]
where $K_{\mathrm{bump}}=I_{\mathrm{tail}}/50(nV_{T})^{2}$. The effect
is shown in figure \ref{fig:Bump}: it makes large errors far from
where the bump is high, but still allows optimisation to find the
top.

\subsection{Attention $A_{st}$ as currents from ``bank''}

The softmax operation (\ref{eq:softmax}) distributes probabilities
among $T$ outputs, exponentially weighted by $T$ inputs. This has
a natural transistor implementation in which a fixed tail current
$I'_{\mathrm{tail}}$ is distributed among $T$ outputs, with each
input $v_{s}$ controlling the gate of one transistor \cite{elfadel1993softmax,sillman2023analog},
as shown in figure \ref{fig:Bank}. We call this circuit ``bank'',
and its output is the following $T$ currents:
\begin{equation}
i^{\mathrm{bank}}_{s}(\vec{v})=I'_{\mathrm{tail}}\frac{e^{v_{s}/\tau}}{\sum^{T}_{b=1}e^{v_{b}/\tau}},\qquad s=1,2,\ldots T,\quad\tau=nV_{T}.\label{eq:current-bank}
\end{equation}
With our NMOS conventions, the fixed tail current is a sink, and so
the output signal is high current flowing into the bank at the $s$
for which $v_{s}$ is largest.

The exact dyadic element has a backward current source obeying (for
case $s=1$, and $T=3$ as in the figure)
\begin{align}
i_{1\to1'} & =\frac{I'_{\mathrm{tail}}}{2\tau}\left[(v_{4}-v_{4}')-C\right]p_{1}\\
\shortintertext{where}C & =(v_{4}-v'_{4})p_{1}+(v_{5}-v'_{5})p_{2}+(v_{6}-v'_{6})p_{3}\\
p_{s} & =e^{v_{s}/\tau}\Big/{\textstyle \sum^{3}_{b=1}e^{v_{b}/\tau}}.\nonumber
\end{align}
After experimenting with various crude approximations, we eventually
use a double circuit which stays quite close to this ideal, as demonstrated
in figure \ref{fig:Bank}. It involves a $2T$-wide forward bank responding
to the mean, $i^{\mathrm{bank}}_{s}(\vec{v}^{\mathrm{mean}})$ with
a voltage divider creating $v^{\mathrm{mean}}_{a}=(v_{a}+v'_{a})/2$;
one VCCS/OTA sensing each output $v_{b}-v'_{b}$ and driving a pair
of reversed banks. Appendix \ref{subsec:Bank} has more details.

\subsection{Current to voltage amplifier}

This softmax-like ``bank'' circuit isn't fed by the ``bump'' scores
directly, for several reasons. Each block accepts voltage but outputs
current, and a high score is a large negative current but needs to
become a positive voltage. In addition, the reverse elements of ``bank''
will disturb the forward output of ``bump''.

The missing link is the current-to-voltage amplifier shown in figure
\ref{fig:Timp}, which inverts with small input and output impedance
$1/G$, and gain controlled by a large resistor. In this arrangement
the VCCS emits current $i=G(0-v_{1})$, all of which flows back through
the resistor, to give $v_{2}-v_{1}=iR$, and flows to the input current
source, thus $i=i_{1\to0}$. Solving, the output is then (defining
$R_{\mathrm{eff}}$):
\begin{equation}
v_{2}=i\:(R-1/G)=i_{1\to0}R_{\mathrm{eff}}.\label{eq:transamp-current}
\end{equation}
In the full circuit, $i_{1}$ comes from bump, and this $v_{2}$ then
controls the bank. (Each output like $v_{2}$ is also pulled up by
a large resistor connected to fixed node. This voltage offset cancels
out of (\ref{eq:current-bank}), but increases the current through
the transistors.)

\begin{figure}
\centering \input{figures/timp.tikz}\captionsetup{width=0.75\textwidth}\caption{Resistor-feedback transimpedance amplifier, using an ideal VCCS (or
operational transconductance amplifier, OTA). Given input current
$i_{1}$, the output voltage is $v_{2}=i_{1}(R-1/G)=i_{1}R_{\mathrm{eff}}$.
Our circuit uses these to convert current outputs from ``bump''
to voltage inputs for ``bank''. }\label{fig:Timp}
\end{figure}

In the doubled circuit, the dyadic potential contributions of the
resistor plus VCCS are $h=[(v_{1}-v_{2})^{2}-(v_{1}'-v_{2}')^{2}]/2R+G(v_{2}-v_{2}')(v_{1}+v_{1}')/2$.
These can be implemented using a similar amplifier in the primed half,
and two floating VCCS: one to pass the error backwards, the other
to correct the forward result, much like the Gilbert cell in section
\ref{subsec:Perfect-voltage-multiplier} above:
\begin{align}
v_{2}' & =R_{\mathrm{eff}}\,i_{1'\to0}\nonumber \\
i_{2'\to2} & =G(v_{1}-v_{1}')/2\label{eq:transamp-adjoint3}\\
i_{1\to1'} & =G(v_{2}-v_{2}')/2.\nonumber
\end{align}

Note aside that this amplifier is very different to the bidirectional
amplifier of Kendall et al. \cite{kendall2020training}. There, within
a single circuit, they inserted an active device which multiplied
all voltages to the right by $A$, and divided all currents to the
left by $A$. This is an active element but their circuits insert
always a line of them, making the whole system gauge-equivalent to
a reciprocal, passive, circuit.

\subsection{Value mixing with differential pairs}

\begin{figure}
\centering \raisebox{5mm}{\input{figures/pair.tikz}}\hspace{5mm}\includegraphics[width=10cm]{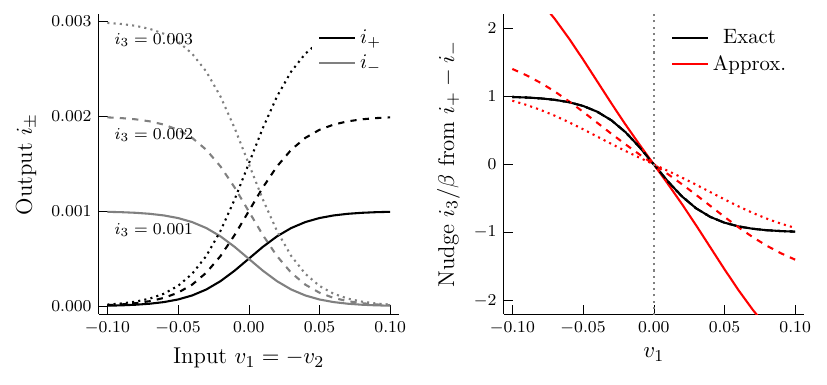}\caption{Differential pair mixer circuit. Here the tail current $i_{3}$ is
an input from bank, while its voltage inputs $\pm x$ will be a token.
It produces a pair of currents $i_{\pm}$, plotted against $x=v_{1}$
for various $i_{3}$. Our reverse circuit provides a gradient signal
only for $i_{3}$; notice that it correctly changes sign at $x=0$.
}\label{fig:Pair}
\end{figure}

We need to use the output of the softmax to create a mixture of the
inputs $X_{\mu s}$ at different times $s$. The differential pair
we use here may be thought of as variant of the ``bank'', taking
$\pm X_{\mu s}$ as the gate voltages. But instead of a fixed tail
current, we divide up the current $i^{\mathrm{bank}}_{s}$, as drawn
in figure \ref{fig:Pair}. Its output is the following pair of currents
(for one time $s$ and one feature $\mu$):
\begin{equation}
i_{\pm}=i^{\mathrm{bank}}_{s}\frac{e^{\pm X_{\mu s}/\tau}}{e^{X_{\mu s}/\tau}+e^{-X_{\mu s}/\tau}}.\label{eq:current-bank1}
\end{equation}
We connect all the $i_{+}$ for different $s$ but the same $\mu$
together, and ground them through a load resistor. And similarly all
the $i_{-}$. Thus a set of $T$ differential pairs gives two voltages
which obey (for one feature $\mu$):
\[
v_{+\mu}-v_{-\mu}=-R_{\mathrm{load}}\sum^{T}_{s=1}i_{s}\tanh(X_{\mu s}/\tau).
\]
This is the desired mixing of different timepoints weighted by the
softmax output. The tanh nonlinearity is unimportant but harmless,
as we control the amplification factor of the input (i.e. the scale
of $X_{\mu s}$ in Volts).

We need $D$ such sets of $T$ differential pairs, one for each feature
$\mu$. In principle all could be driven by the same bank output.
But each draws some current, and the naiive wiring means they interfere
with each other. Our solution at present is to simply have $D$ equivalent
bank circuits driving them.

For the gradient of the differential pair mixer, we again use Gilbert
cells to read $\Delta v_{44'}$ and $\Delta v_{55'}$ from the output,
and send current back towards ``bank'' via $i_{3}$ :
\[
i_{3\to3'}=K_{\mathrm{pair}}(v_{1}-v_{2})(v_{4}'-v_{4})+K_{\mathrm{pair}}(v_{1}-v_{2})(v_{5}-v_{5}')
\]
with $K_{\mathrm{pair}}=I_{\mathrm{ref}}/8nV^{2}_{T}$ involving a
reference tail current at which the slope matches. This current captures
the important feature of the true gradient that it changes sign when
$x$ does -- see figure \ref{fig:Pair}. Since there is only one
layer we do not supply an error signal towards the gates (i.e. towards
$X_{\mu s}$).

The final output comes from another resistor array, $D_{\mathrm{out}}\times2D$,
implementing the value weights, roughly
\[
y_{\mu}=\sum_{\pm\nu}W^{\mathrm{value}}_{\mu,\pm\nu}v_{\pm\nu},\qquad\mu=1,2\ldots D_{\mathrm{out}}.
\]

\subsection{Limitations}

The circuit described here is an idealised sketch, and not a realistic
engineering design. The circuit elements we employ have all been built,
with approximately the functional properties shown, but they have
limitations which our simulations do not model. Even in this sketch,
there are many parameters tuned --- tail currents, fixed resistances,
etc --- which help match the different layers. We are quite certain
there are designs which are more realistic, more elegant, and use
fewer components.

Table \ref{tab:Component-counts} lists the components needed for
each of the layers described. Here we take $D=12$, $T=5$, and $D_{\mathrm{out}}=6$,
which is the size of the model trained to perform in-context learning
in the next section. We observe that the double circuit is larger
by an $\mathcal{O}(1)$ factor, rather than growing with the number
of parameters --- the same property as reverse-mode automatic differentiation
\cite{speelpenning1980compiling}.

This model has 444 learned parameters, which is an order of magnitude
larger than the breadboard self-learning circuits of \cite{dillavou2026understanding}.
But still smaller than the crossbar arrays of \cite{yi2022activitydifference}
with $64^{2}$ TaOx cells, used there to simulate an MLP with 1000
parameters.

\begin{table}
\centering \footnotesize %
\begin{tabular}{l|l|l|r@{\extracolsep{0pt}.}l}
Layer & Circuit ($N=497$ nodes) & Double circuit ($N=1198$) & \multicolumn{2}{c}{Parameters}\tabularnewline
\hline
\hline
Key (array) & 120 resistors, 60 VS & 240 resistors, 120 VS & \multicolumn{2}{c}{12}\tabularnewline
Query (array) & 288 resistors, 12 VS & 576 resistors, 24 VS & \multicolumn{2}{c}{288}\tabularnewline
Scores (bump) & 240 transistors, 60 CS & 480 transistors, 120 CS, 240 multipliers & \multicolumn{2}{c}{0}\tabularnewline
Amplifier & 5 VCCS, 10 resistors & 20 VCCS, 20 resistors & \multicolumn{2}{c}{0}\tabularnewline
Attention (bank) & 60 transistors, 24 resistors, 12 CS & 480 transistors, 168 resistors, 36 CS & \multicolumn{2}{c}{0}\tabularnewline
Mixing (pairs) & 120 transistors & 240 transistors, 120 multipliers & \multicolumn{2}{c}{0}\tabularnewline
Value (array) & 144 resistors & 288 resistors & \multicolumn{2}{c}{144}\tabularnewline
Loss (LCE) & --- & 6 transistors, 12 VCCS, 24 resistors, 1 CS & \multicolumn{2}{c}{0}\tabularnewline
\hline
Total & 1155 components & 3217 components & \multicolumn{2}{c}{444}\tabularnewline
\end{tabular}

\caption{Component counts for the training task. Here $D=12$, $T=5$, and
$D_{\mathrm{out}}=6$. VCCS is a floating voltage-controlled current
source, CS is a fixed current source, VS is a floating voltage-difference
device, multiplier is an ideal Gilbert cell. Counts do not include
the input and target voltage sources. The rightmost column is the
number of independently trained resistances, which is unchanged by
doubling. }\label{tab:Component-counts}
\end{table}

\section{Minimal in-context learning task}\label{sec:Minimal-ICL-task}

In-context learning (ICL) is the name given to some kinds of generalisation
at which transformers excel \cite{brown2020language}. It is often
contrasted to memorisation \cite{chan2022datadistributional,reddy2024mechanistic},
and can be illustrated by examples like this:
\begin{itemize}
\item Q: ``Who was the first man on the moon?'' A: ``Neil Armstrong''.
Clearly his name had to be memorised, stored in the trained weights.
\item Q: ``Anna \& Grzegorz walked up the hill this morning, and Anna came
down at lunchtime. Who is still up there?'' A: ``Grzegorz!''. Here
the question in the second sentence doesn't have a unique answer,
but (having learned the rules of English) it's easy to solve by copying
letters from the first sentence, even if the name is unfamiliar.
\end{itemize}
The transformer here receives the question encoded such that each
letter is a column of the input matrix $X$, and gives probabilities
for the next character. Repeated sampling generates the answer string.

The associative recall task we use here has a much simpler grammar
than English, and consists of strings like ``a1 e5 c3 a?'', where
the task is to predict the last character \cite{ba2016using,chan2022datadistributional,reddy2024mechanistic}.
Every string in the training set has the answer appearing earlier,
so that this can be solved by looking backwards, finding the matching
token ``a'', and copying the number after it, ``1'', as the answer.
If all the strings use a fixed letter-number relationship (like a$\Rightarrow1$,
b$\Rightarrow2,\ldots$) then this can also be solved by memorising
this relationship. The network will tend to learn just one solution,
and the training data determines which is easier \cite{chan2022datadistributional,reddy2024mechanistic}
--- see appendix \ref{sec:Training-for-memorisation} for more details.
Here we use a large number of letters to ensure that ICL is preferred.

\begin{figure}
\centering \includegraphics[width=15cm]{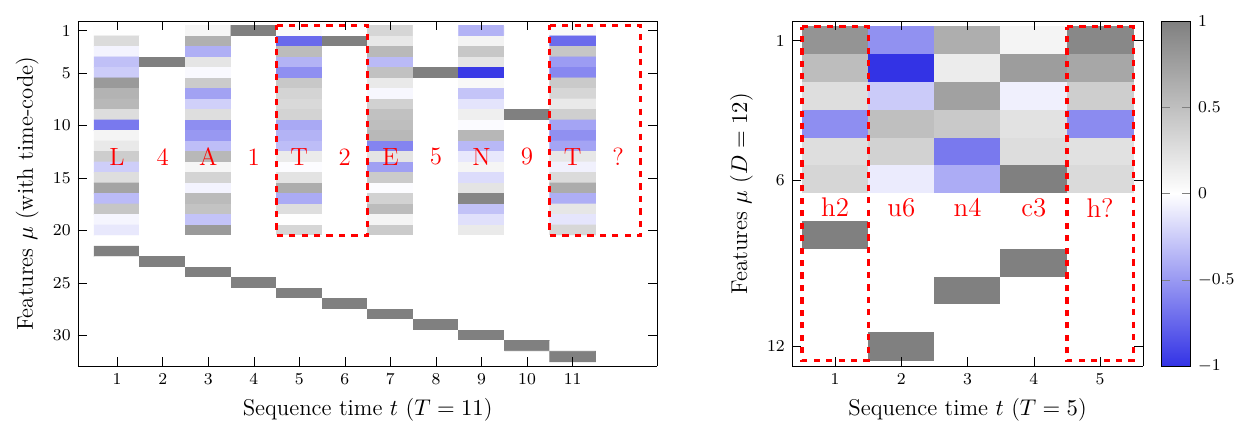}

\caption{Input matrices $X_{\mu t}$ for the in-context learning (ICL) task,
adapted from \cite{reddy2024mechanistic,nguyen2025differential}.
Left, each character of string ``L4 A1 T2 E5 N9 T'' is a 20-dimensional
embedding vector, with letters (random) and numbers (``one-hot'')
alternating in time. They are accompanied by a simple time-code, in
which the nonzero $\mu$ is incremented from a random start. This
input needs a two-layer transformer, and the first layer's function
is to pay attention to time $t-1$ to encode letter and number together,
giving a result like the matrix shown at right. Right, the 6-dimensional
embedding vectors for a letter and a number are concatenated to one
12-dimensional vector, for each pair in ``h2 u6 n4 c3 h?''. The
desired output is $y^{\mathrm{target}}=[0,1,0,0,0,0]$, representing
``2''. This task can be learned by a one-layer transformer, which
must pay attention to time $t=1$ where the vector representing the
query ``h'' appears. The $D=12,T=5$ case shown is precisely the
task studied in this paper.}\label{fig:Data-ICL}
\end{figure}

Input matrices for this task are shown in figure \ref{fig:Data-ICL}.
If these letters and numbers alternate (i.e. occupy even and odd columns
of the input matrix) then this task needs two attention layers, or
one induction head. But if the vectors representing letters and numbers
are concatenated (so that ``a1'' is one vector with entries for
``a'' followed by entries for ``1'') as in \cite{nguyen2025differential}
then a single attention layer is sufficient. This is the simple case
we study, using the electrical circuit introduced above.

\begin{figure}
\centering \includegraphics[width=13cm]{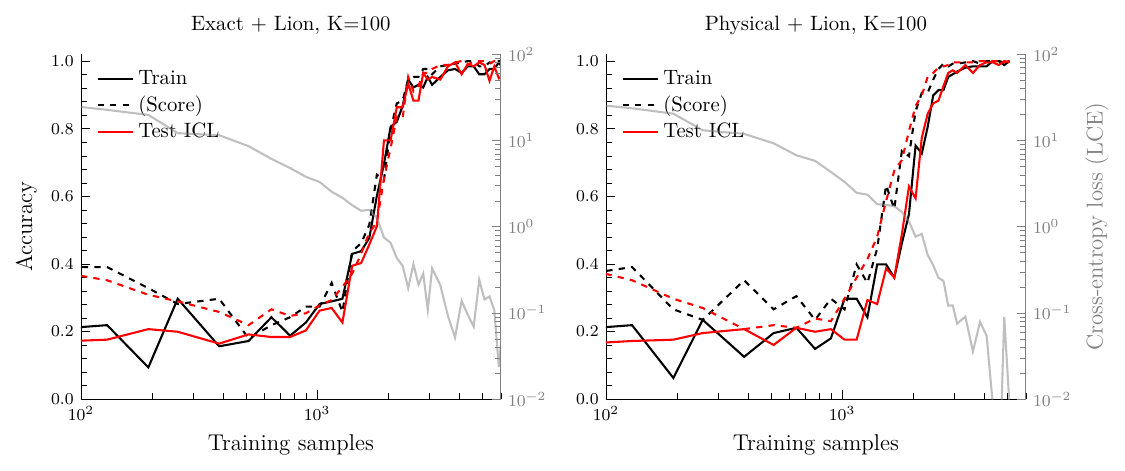}

\caption{Successful training on the ICL task, using either exact gradients
(dyadic circuit) at left, or approximate versions at right. We track
accuracy both at the final output, and also by looking at whether
the score is highest at the time $s$ which contains the solution.
 A random output will have accuracy $1/6=1/D_{\mathrm{out}}$, but
simply copying a random part of the input (without attention) will
give accuracy $1/4$. All use Lion with $\eta=3\times10^{-9}$S and
$\beta_{1}=\beta_{2}=0.99$. The training data uses $K=100$ distinct
``letter''s encoded as unit vectors in $\mathbb{R}^{6}$ to make
memorisation difficult; see appendix \ref{sec:Training-for-memorisation}
for other cases.}\label{fig:Training-ICL}
\end{figure}

For training to succeed, we find it essential to use both LCE loss
(not MSE), and to use an adaptive optimiser such as Adam or Lion (not
gradient descent). Figure \ref{fig:Training-ICL} shows training progress,
using either exact gradients, or the crude approximate gradient components
introduced above. We choose the initial resistors randomly, log-uniform
in $10^{3}\Omega\leq r_{ab}\leq10^{8}\Omega$, and constrain them
during training to this same range.

To test ICL accuracy, we use test cases like ``\textgreek{β}2 \textgreek{ζ}6
\textgreek{ε}5 \textgreek{ζ}?''. That is, we generate new embedding
vectors outside the training set, and check that the circuit is able
to match these and route their companion numbers correctly \cite{reddy2024mechanistic}.
Figure \ref{fig:Training-ICL} uses a fixed test set of 256 examples,
created using 200 new random embedding vectors.

\section{Conclusion}

In this paper we have shown how to construct a transformer-like architecture
which allows for physical learning. In some sense this is an engineering
project --- we've translated an architecture known to have certain
learning properties to a different substrate, and translated also
the reverse pass and the training law.  We brought together a range
of ingredients to achieve this. But the idea of translating a known
architecture to allow physical learning is familiar, except that it's
mostly been done for 20th century architectures: multi-layer perceptrons
(MLPs) \cite{kendall2020training} and convolutional nets \cite{laborieux2020scaling}.

However, we'd like to claim a slightly grander view as motivation.
Engineering unconventional hardware isn't the only motivation for
thinking about physical learning; instead, we hope that learning about
learning in many contexts may ultimately teach us about how biology
learns. While MLPs were motivated by networks of biological neurons,
the extreme un-realism of training them by back-propagation has been
a longstanding flaw in this analogy, famously noted by Crick \cite{crick1989recent}.
In this light our work follows Lillicrap and Liao \cite{lillicrap2016random,liao2016how}
in asking how crude a variant of exact back-propagation might still
work. But we add also a more difficult learning task.

The task we train the system to perform, a form of in-context learning
(ICL), is quite different from any considered within physical learning.
While feed-forward networks (like MLPs) have no trouble learning classification
tasks like MNIST digits, these associative recall tasks appear to
be very difficult for them to learn. Such tasks are at the core of
our best understanding of what makes transformers so different from
earlier models \cite{olsson2022context,arora2023zoology}. Of course
we don't think the brain is literally a transformer, but we do think
that our ease of solving such tasks ought to draw attention towards
architectures which are able to do so. Toy ICL tasks are, perhaps,
a useful diagnostic of whether a small system has post-MLP learning
abilities.

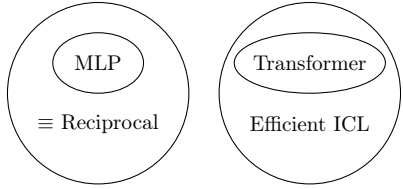
\begin{figure}
\centering \begin{tikzpicture}[scale=0.4, every node/.style={scale=0.8}]
  \draw (-4,0) circle (3);
  \draw (-4,1) ellipse (1.5 and 1);
  \node at (-4,1)    {MLP};
  \node at (-4,-1) {$\equiv$ Reciprocal};

  \draw (3,0) circle (3);
  \draw (3,1) ellipse (2.5 and 1);
  \node at (3,1)    {Transformer};
  \node at (3,-1) {Efficient ICL};
\end{tikzpicture} \captionsetup{width=0.75\textwidth}

\caption{Our conjectured Venn diagram for physical learning systems. A digital
MLP (with relu) is not a reciprocal system, but is equivalent to network
of resistors and diodes. The transformer presented here has explicitly
non-reciprocal interactions, and needs our doubled formalism to enable
physical learning. It is good at learning the ICL task. }\label{fig:Venn-diagram}

\end{figure}

Physical learning has mostly concerned itself with physical systems
of reciprocal pairwise interactions --- the resistors and diodes
of section \ref{sec:Circuits-adapted-for-learning}, but also spring
systems \cite{pashine2019directed,stern2020supervised} and pipe networks
\cite{stern2021supervised,anisetti2023learning}, chemical interactions
\cite{poole2025autonomous} and spin systems \cite{debos2026multimodal}.
Our conjecture is that no such reciprocal system can efficiently learn
ICL tasks --- as sketched in figure \ref{fig:Venn-diagram}. It's
hard to prove statements about inductive bias, and usually requires
specifying many architectural details, as for instance in \cite{li2021whyconvolutional,malach2021computational}.
We investigate numerically in appendix \ref{sec:Training-for-memorisation};
see also \cite{tong2025mlps} for related work on ICL in digital neural
networks. What the field of machine learning has been good at proving
are results about what can be expressed by different architectures.
MLPs of sufficient width can express any function \cite{cybenko1989approximation,leshno1993multilayer},
and the same is true of their circuit relatives \cite{scellier2024universal}.
Results about deep MLPs being exponentially more expressive \cite{poole2016exponential}
can surely also be reproduced. But expressivity is an upper bound
on what can be learned, and often an extremely loose one.

While non-reciprocal circuits cannot in general be trained by blindly
applying equilibrium propagation, a larger circuit always exists with
the same forward properties, and explicit reverse propagation \cite{scurria2026equilibrium}.
However, such dyadic circuits will in general be much harder to physically
realise, as they require components with arbitrary new current laws.
Our simpler doubled circuits use realistic components to transmit
approximate gradient information. The number of reverse elements is
an order-1 factor larger than the number of original forward components,
parallel to $\mathcal{O}(1)$ complexity results for digital networks
\cite{speelpenning1980compiling}.

\subsection{The broader landscape of unconventional learning}

The physical learning systems we study here are part of a broader
revival of interest in analogue computing, aimed at machine learning
tasks presently performed by digital computers. Two major motivations
are energy and time: for example, a circuit to multiply two voltages
needs far fewer transistors than the digital one to multiply two floating-point
numbers \cite{sarpeshkar1998analog}, and need only settle to its
operating point once, instead of once per clock cycle. Most such systems
are studied without the constraint that analogue physics transports
both the ``forward'' or inference signal and the ``backward''
error signal, which we took as the definition of physical learning.

At the opposite extreme, the training can take place in simulation,
before the physical system is constructed. The task we studied above
is in fact so simple that we can just write down the weights required
by hand, and could easily hard-wire the circuit. But in general this
process is very sensitive to small defects, and various hybrid methods
aim to train the system to compensate, for instance by measuring the
physical output of each layer and simulating only a layer-wise gradient
\cite{wright2022deep,nest2024training}. These ideas have been applied
to a wide range of physical substrates \cite{momeni2025training}.
Most of them are, we believe, in the same learning class as MLPs.
That is, most are essentially feed-forward networks without the comparison
needed for associative learning. But there is no obvious bar to using
hybrid methods on systems like the transistor circuit described here.

Hopfield networks \cite{hopfield1982neural} are another class of
unconventional learning systems, in which memories are stored as minima
of an energy function. The traditional test is how well they can reconstruct
a memory from corrupted input --- that is, the input is the initial
value of a dynamical state, which then evolves back to the stored
memory. This is a task quite unlike either digit classification or
our ICL. Recently, a variety of Hopfield-inspired networks have been
constructed which aim to capture more of the transformer's behaviour
\cite{ramsauer2021hopfield,hoover2023energy,dehmamy2026nrgpt,hoier2026training}.
These are often described as energy-based models, but energy plays
a different role to its meaning in steady-state reciprocal circuits
(like figure \ref{fig:Single-MLP}): in \cite{ramsauer2021hopfield,hoover2023energy}
the application of a transformer layer is modelled as one step downhill
in an energy, and in \cite{dehmamy2026nrgpt} there is a different
energy per token. That is if $x_{t}$ is the token for sequence time
$t$, its evolution through one step downhill is
\[
x^{(\mathrm{next})}_{t}=x_{t}-\eta\nabla E(x_{t}|\{x_{s},s<t\})
\]
which allows non-reciprocal interactions between $x_{s}$ and $x_{t}$.
These models are trained by finding exact gradients of the forward
process, and by construction they behave very much like transformers.
The exception is the CET model of \cite{hoier2026training}, where
the output truly is the result of a minimisation over the output and
hidden degrees of freedom, holding the input $X$ fixed:
\[
y=\argmax_{y,Z}E_{\theta}(X,y,Z).
\]
Here the energy contains terms like $\sum_{t}\log\sum_{s}\exp\gamma S_{st}$
using an attention score similar to (\ref{eq:attention}). They train
this model by equilibrium propagation, and show that it can learn
to reconstruct corrupted images. In our adaptation, appendix \ref{subsec:CET},
it learns memorisation but not to perform ICL --- in line with our
conjecture about the need for non-reciprocal interactions for this
task.

\subsection*{Acknowledgements}

We thank Pranav Kantroo and Matthew Leighton for helpful comments
on a draft.

This work was aided by OpenAI's ChatGPT for Academic Researchers program.
In particular, the simulation code and the tuning of all details of
this circuit would have been a huge project before. None of the text
was written by AI.

This work was performed in part at the Aspen Center for Physics, which
is supported by NSF grant PHY-2210452.

\bibliographystyle{my-JHEP-5}
\bibliography{/Users/me/Documents/Current-Work/0000-zotero-export/My-Library}

\newpage

\appendix

\renewcommand{\thefigure}{S\arabic{figure}}
\setcounter{figure}{0}

\renewcommand{\thetable}{S\arabic{table}}
\setcounter{table}{0}

\renewcommand{\thesection}{\Alph{section}}
\renewcommand{\thesubsection}{\Alph{section}.\arabic{subsection}}

\counterwithin{equation}{section}
\renewcommand{\theequation}{\thesection\arabic{equation}}

\section*{\centering\LARGE Appendix}


\begin{figure}[H]
\centering \includegraphics[width=9cm]{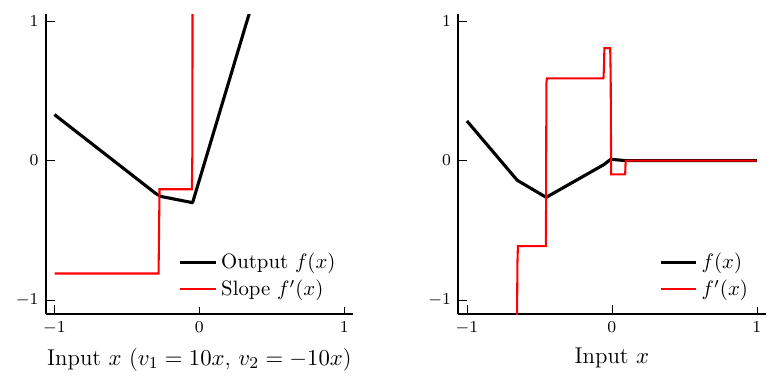}\captionsetup{width=0.75\textwidth}

\caption{Behaviour of the resistor-diode circuit shown in figure \ref{fig:Single-MLP},
with exactly three hidden nodes. Left, with randomly chosen resistors
(log-uniform in $0.1\Omega$ to $10\Omega$) there are typically three
nonzero slopes, and two or three kinks. Right, resistors can be found
such that this has five nonzero slopes, and five kinks. This would
be impossible for a real MLP with three hidden nodes, equation (\ref{eq:MLP}),
and results from current through the output resistors from one hidden
node perturbing other hidden nodes.}\label{fig:Five-kinks}
\end{figure}

\section{Deriving current rules for bidirectional information flow}\label{sec:Deriving-current-rules}

Here we show explicitly how to construct dyadic circuit elements using
the formalism of Scurria et al. \cite{scurria2026equilibrium}, and
then discuss the approximations made for the circuit blocks in section
\ref{sec:Transformer-circuit}. The dyadic construction of \cite{hoier2024dyadic}
is similar; it uses for instance $+\beta L(z)+\beta L(z')$ in place
of $2\beta L(\frac{z+z'}{2})$ which produces deviations at $\mathcal{O}(\beta^{2})$.
This is also related to Galley's doubled-history variational principle
\cite{galley2013classical} with some important differences, including
that the final physical solution there turns off the difference, $z-z'\to0$,
while \cite{scurria2026equilibrium} keeps it.

Consider a dynamical system $\dot{x}_{a}=F_{a}(x,\theta)$ where $x\in\mathbb{R}^{N'}$
is the state and $\theta$ some parameters. Conservative systems are
those where $F_{a}(x,\theta)=-\partial E(x,\theta)/\partial x_{a}$,
and the standard setting for equilibrium propagation considers the
interplay between minimising this energy and a loss function $L(x)$,
as for example in \cite{stern2025physical}. Finding stationary states
$x^{\beta}$ which minimise the augmented energy $E(x,\theta)+\beta L(x)$
for different $\beta$ allows us to find gradients
\[
\frac{dL(x^{0})}{d\theta}=\frac{d^{2}E(x^{\beta},\theta)}{d\beta\,\partial\theta}=\lim_{\beta\to0}\frac{1}{2\beta}\left[\frac{\partial E(x^{\beta},\theta)}{\partial\theta}-\frac{\partial E(x^{-\beta},\theta)}{\partial\theta}\right]
\]
and hence to minimise $L(\theta)$ by gradient descent. Locally, an
energy function exists iff $\partial F_{a}/\partial x_{b}=\partial F_{b}/\partial x_{a}$,
that is, the Jacobian is symmetric, or the system has reciprocal interactions.

For non-conservative systems, \cite{scurria2026equilibrium} introduce
a dyadic system which has two copies of the original state, $z,z'\in\mathbb{R}^{N'}$,
and a dyadic potential defined
\begin{equation}
H(z,z',\theta)=-\sum_{a}(z_{a}-z'_{a})\:F_{a}\Big(\frac{z+z'}{2},\theta\Big).\label{eq:scurria-potential}
\end{equation}
The new dynamics are then $\dot{z}_{a}=-\partial H(z,z',\theta)/\partial z_{a}$
and $\dot{z}'_{a}=+\partial H(z,z',\theta)/\partial z'_{a}$. For
a linear reciprocal system (such as the resistor networks below),
this potential is $H(z,z',\theta)=E(z,\theta)-E(z',\theta)$, thus
we have two identical copies with no interactions, and a stationary
state at a saddle of $H$. To allow equilibrium propagation, they
consider an augmented potential $H_{T}=H(z,z',\theta)+2\beta L(\frac{z+z'}{2})$.
A stationary state then obeys
\[
0=\frac{\partial H_{T}}{\partial z_{a}}\pm\frac{\partial H_{T}}{\partial z'_{a}}=\begin{cases}
-2F_{a}(\frac{z+z'}{2},\theta), & -\\
-\sum_{b}d_{b}\frac{\partial F_{b}(m,\theta)}{\partial m_{a}}+2\beta\frac{\partial L(m)}{\partial m_{a}}, & +.
\end{cases}
\]
Thus the mean $m=(z+z')/2$ solves the original equation, while the
difference $d=(z-z')$ is an error signal, proportional to $\beta$.
They show that, as a result, the gradient is
\begin{equation}
\frac{dL}{d\theta}=\frac{1}{2\beta}\frac{\partial H(z^{\beta},z^{\prime\beta},\theta)}{\partial\theta}.\label{eq:scurria-gradient}
\end{equation}
This is exact, with no limit $\beta\to0$ required.

To apply this to electrical circuits, we take $F_{a}(x,\theta)$ to
be the total current flowing into free node $a$, and $x_{a}=v_{a}$
to be the voltage at this node. We consider only DC circuits, so there
is no time-dynamics $\dot{x}$, but the relations above concerning
stationary states still hold. At the free nodes, $F_{a}(x,\theta)=0$
is precisely the KCL. Besides the free nodes, our circuits include
several at fixed voltages (including $v_{0}=0$, ground).

Given a circuit with $N'$ free nodes described by $F(x,\theta)$,
the above procedure always gives a circuit with $2N'$ free nodes
described by $H(z,z',\theta)$, with $v_{a}=z_{a}$ and $v_{a'}=z'_{a}$.
This dyadic circuit has the same parameters $\theta$, perhaps the
conductances of $M$ variable resistors. Further, suppose that we
decompose the circuit into components
\[
F(x,\theta)=\sum_{q}f^{(q)}(x,\theta)
\]
where each contribution $f^{(q)}$ depends on just a few node voltages,
and influences only a few currents, together say $\{a,b,c,\ldots\}$.
Then clearly there is a similar decomposition in the dyadic system:
\[
H(z,z',\theta)=\sum_{q}h^{(q)}(z,z',\theta)
\]
where $h^{(q)}$ depends only on nodes $\{a,a',b,b',c,c',\ldots\}$.
Both $f^{(q)}$ and $h^{(q)}$ depend on the same subset of the parameters.
In general $h^{(q)}$ will describe a block of physical components.
The current laws of this dyadic block may be read from
\begin{equation}
i_{0\to a}=-\frac{\partial h^{(q)}(z,z',\theta)}{\partial z_{a}},\qquad i_{0\to a'}=+\frac{\partial h^{(q)}(z,z',\theta)}{\partial z'_{a}}\label{eq:scurria-currents}
\end{equation}
which follow from dynamics for $\dot{z},\dot{z}'$ above.

In this section we stay closer to the notation of \cite{scurria2026equilibrium},
and in particular use $z,z'$ for node voltages in the double/dyadic
systems (while the main text uses $v,v'$.)

\subsection{Resistor}

As a warm-up example, we consider the simplest possible circuit element
with
\[
i_{a\to b}=g_{ab}(v_{a}-v_{b}).
\]
The currents are $f^{\mathrm{res}}_{a}(x)=-f^{\mathrm{res}}_{b}(x)=-g_{ab}(v_{a}-v_{b})$,
which agree with energy (really half the power dissipated):
\[
E^{\mathrm{res}}(x)=(g_{ab}/2)(v_{a}-v_{b})^{2}.
\]
 The contribution to the dyadic potential is
\begin{align*}
h^{\mathrm{res}} & =(z_{a}-z_{a}')g_{ab}\frac{z_{a}-z_{b}+z'_{a}-z'_{b}}{2}-(z_{b}-z_{b}')g_{ab}\frac{z_{a}-z_{b}+z'_{a}-z'_{b}}{2}\\
 & =(g_{ab}/2)(z_{a}-z_{b})^{2}-(g_{ab}/2)(z'_{a}-z'_{b})^{2}.
\end{align*}
As advertised, this is the difference of two single-copy potentials,
$E^{\mathrm{res}}(z)-E^{\mathrm{res}}(z')$. The resulting currents
are
\begin{align*}
i_{0\to a} & =-g_{ab}(z_{a}-z_{b}) & i_{0\to a'} & =-g_{ab}(z'_{a}-z'_{b})\\
i_{0\to b} & =+g_{ab}(z_{a}-z_{b}) & i_{0\to b'} & =+g_{ab}(z'_{a}-z'_{b})
\end{align*}
which we can combine into two resistors, both sharing the same conductance
$g_{ab}$:
\begin{align*}
i_{a\to b} & =g_{ab}(z_{a}-z_{b}) & i_{a'\to b'} & =g_{ab}(z'_{a}-z'_{b}).
\end{align*}

\subsection{Gilbert cell --- figure \ref{fig:Denver}}\label{subsec:Gilbert}

Recall that this is a 6-terminal object with current law:
\[
i_{e\to f}=K(v_{a}-v_{b})(v_{c}-v_{d}).\tag{\ref{eq:current-Gilbert}}
\]
In the circuit of figure \ref{fig:Denver}, half the terminals are
grounded, leaving just
\[
i_{0\to5}=Kv_{3}v_{4}.\tag{\ref{eq:current-Gilbert0}}
\]
The only current is into node 5, thus only $f^{\mathrm{gilbert}}_{5}(x)=Kv_{3}v_{4}$
matters here. The dyadic potential contribution is
\[
h^{\mathrm{gilbert}}(z,z')=-(z_{5}-z_{5}')\frac{K}{4}(z_{3}+z_{3}')(z_{4}+z_{4}').
\]
The currents which dyadic components must deliver are then read from
(\ref{eq:scurria-currents}):
\begin{align}
i_{0\to5} & =\frac{K}{4}(z_{3}+z_{3}')(z_{4}+z'_{4}) & i_{0\to5'} & =\frac{K}{4}(z_{3}+z_{3}')(z_{4}+z'_{4}) &  & \text{forward}\nonumber \\
i_{0\to4} & =\frac{K}{4}(z_{3}+z_{3}')(z_{5}-z'_{5}) & i_{0\to4'} & =-\frac{K}{4}(z_{3}+z_{3}')(z_{5}-z'_{5}) &  & \text{backward}\\
i_{0\to3} & =\frac{K}{4}(z_{4}+z_{4}')(z_{5}-z'_{5}) & i_{0\to3'} & =-\frac{K}{4}(z_{4}+z_{4}')(z_{5}-z'_{5}).\nonumber
\end{align}
Here we no longer assume that node $3$ is fixed. The backward currents
can be implemented by four floating Gilbert cells:
\[
\begin{aligned}i_{4'\to4} & =\frac{K}{4}z_{3}(z_{5}-z'_{5})+\frac{K}{4}z_{3}'(z_{5}-z'_{5})\\
i_{3'\to3} & =\frac{K}{4}z_{4}(z_{5}-z'_{5})+\frac{K}{4}z_{4}'(z_{5}-z'_{5}).
\end{aligned}
\]
For the forward currents, expanding $(z_{3}+z'_{3})(z_{4}+z'_{4})=z_{3}z_{4}+z_{3}z'_{4}+z'_{3}z_{4}+z'_{3}z'_{4}$
gives a set of eight forward multipliers. Our approximation is to
replace these by four --- the original, the obvious double, and two
correction terms:
\[
\begin{aligned}i_{0\to5} & =Kz_{3}z_{4}\\
i_{0\to5'} & =Kz'_{3}z'_{4}\\
i_{5\to5'} & =\frac{K}{2}z_{3}z_{4}-\frac{K}{2}z'_{3}z'_{4}.
\end{aligned}
\]
This makes an error $(K/4)(z_{3}-z'_{3})(z_{4}-z'_{4})$, which is
$\mathcal{O}(\beta^{2})$. For the case in section \ref{subsec:Perfect-voltage-multiplier}
in the main text, $z_{3}=z'_{3}$ so it becomes exact (and the two
correction terms become one).

\subsection{Diode}\label{subsec:Diode}

Consider a diode with small finite resistance $1/g$ when it is turned
on, thus current law
\[
i_{a\to b}=g\max(v_{a}-v_{b},0)
\]
and write $\Delta v_{ab}=v_{a}-v_{b}$, and $E(\Delta v_{ab})=(i_{a\to b})^{2}/2g$.
This has dyadic energy contribution
\[
h^{\mathrm{diode}}(z,z')=(\Delta z_{ab}-\Delta z'_{ab})\:g\max\Big(\frac{\Delta z_{ab}+\Delta z'_{ab}}{2},0\Big).
\]
We can think of this as a 4-terminal component which responds to $m=(\Delta z_{ab}+\Delta z'_{ab})/2$
by changing its resistance on both paths $a\to b$ and $a'\to b'$
together:
\begin{align*}
i_{a\to b} & =\begin{cases}
g\Delta z_{ab}, & m>0\\
0, & m<0
\end{cases}\\
i_{a'\to b'} & =\begin{cases}
g\Delta z'_{ab}, & m>0\\
0, & m<0.
\end{cases}
\end{align*}

The obvious simpler component is two independent diodes, which have
instead
\[
h_{\mathrm{indep}}=E(\Delta z_{ab})-E(\Delta z'_{ab}).
\]
This differs when one diode is on and the other is off, and can introduce
a difference the same scale as the error signal. This is why we write
$\approx$ in equation (\ref{eq:gradient-double}) without further
qualification.

\subsection{MOSFET}\label{subsec:MOSFET}

Recall that our device has
\[
i_{d\to s}=\mathrm{mosfet}(v_{g},v_{s},v_{d})=I_{0}e^{(v_{g}-V_{\mathrm{th}})/nV_{T}}\left[e^{-v_{s}/V_{T}}-e^{-v_{d}/V_{T}}\right].\tag{\ref{eq:current-mosfet}}
\]
Thus $f^{\mathrm{mosfet}}_{s}(x)=-f^{\mathrm{mosfet}}_{d}(x)=\mathrm{mosfet}(v_{g},v_{s},v_{d})$,
and
\[
h^{\mathrm{mosfet}}(z,z')=(z_{d}-z'_{d}-z_{s}+z_{s}')\mathrm{mosfet}(\frac{z_{g}+z'_{g}}{2},\frac{z_{s}+z'_{s}}{2},\frac{z_{d}+z'_{d}}{2})
\]
from which we derive the 6 currents out of the dyadic MOSFET device.
The important thing is that the dyadic component has $i_{0\to g}$,
current flows back from the gate.

\subsection{Bump --- figure \ref{fig:Bump}}\label{subsec:Bump}

This block has
\[
i_{5\to0}=\frac{I_{\mathrm{tail}}}{1+4\cosh^{2}\frac{v_{1}-v_{2}}{2\tau}}=\mathrm{bump}(v_{1}-v_{2}).
\]
with $\tau,I_{\mathrm{tail}}$ fixed \& positive (and nodes 3,4 internal).
This ``output'' current flows into the device, and $I_{\mathrm{tail}}$
is current flowing out to ground. (Current is drawn from 3,4 which
also connect to ground.)

Since current only affects one node 5, we need only $f^{\mathrm{bump}}_{5}(x)$,
and get this contribution:
\[
h^{\mathrm{bump}}(z,z')=(z_{5}-z_{5}')\mathrm{bump}\Big(\frac{z_{1}+z_{1}'}{2}-\frac{z_{2}+z_{2}'}{2}\Big)
\]
and then we can compute six current laws like
\begin{align*}
i_{5\to0} & =\frac{\partial h^{\mathrm{bump}}}{\partial z_{5}}=\mathrm{bump}\Big(\frac{z_{1}+z_{1}'}{2}-\frac{z_{2}+z_{2}'}{2}\Big)\\
i_{0\to1} & =-\frac{\partial h^{\mathrm{bump}}}{\partial z_{1}}=-(z_{5}-z_{5}')\frac{1}{2}\mathrm{bump}'(...).
\end{align*}
The physical approximation we use is this, a linearisation about the
top of the bump, $z_{1}+z_{1}'=z_{2}+z_{2}'$:
\begin{align*}
i_{5\to0} & \approx\mathrm{bump}(z_{1}-z_{2})\\
i_{5'\to0} & \approx\mathrm{bump}(z'_{1}-z'_{2})\\
i_{1\to1'}=-i_{2\to2'} & \approx-\frac{I_{\mathrm{tail}}}{50\tau^{2}}(z_{5}-z_{5}')(z_{1}+z'_{1}-z_{2}-z'_{2})
\end{align*}

\subsection{Bank --- figure \ref{fig:Bank}}\label{subsec:Bank}

This block has $M$ inputs and $M$ outputs, with:
\[
i_{a+M\to0}=\frac{I_{\mathrm{tail}}e^{v_{a}/\tau}}{Z}=I_{\mathrm{tail}}p_{a},\qquad Z=\sum^{M}_{b=1}e^{v_{b}/\tau}
\]
defining $p_{a}=e^{v_{a}/\tau}/Z$. Let me take $M=3$ to match the
figure. This draws current from each node $a+3$ with $f^{\mathrm{bank}}_{a+3}(x)=-I_{\mathrm{tail}}p_{a}$,
hence the dyadic energy contribution is
\[
h^{\mathrm{bank}}(z,z')=-\sum_{a=1,2,3}(z_{a+3}-z'_{a+3})f_{a+3}(\frac{z+z'}{2}).
\]
The forward currents are like this
\begin{align*}
i_{4\to0} & =i_{4'\to0}=I_{\mathrm{tail}}p_{1}\\
i_{5\to0} & =i_{5'\to0}=I_{\mathrm{tail}}p_{2}\\
i_{6\to0} & =i_{6'\to0}=I_{\mathrm{tail}}p_{3}
\end{align*}
with $p_{a}$ evaluated at $(z+z')/2$. We approximate these by $i_{4\to0}\approx I_{\mathrm{tail}}p_{1}$
evaluated at $z$, and $i_{4'\to0}\approx I_{\mathrm{tail}}p_{1}$
evaluated at $z'$, etc.

\begin{figure}
\centering \input{figures/bank.tikz}\hspace{1cm}\input{figures/bank_meangadget.tikz}\caption{Bank has inputs $1,2,3$ and outputs $4,5,6$. At right we draw its
doubled circuit, with the output nodes in the middle, and input nodes
drawn twice. Forwards, a 6-wide bank responds to $m_{a}=(v_{a}+v'_{a})/2$,
explicitly invariant to input differences $(v_{a}-v'_{a})$, and with
independently created $i_{4\to0}$ and $i_{4'\to0}$ to reduce interference.
Backward, two 3-wide banks each respond to the same $m_{a}$, shifted
slightly. }\label{fig:Bank-meangadget}
\end{figure}
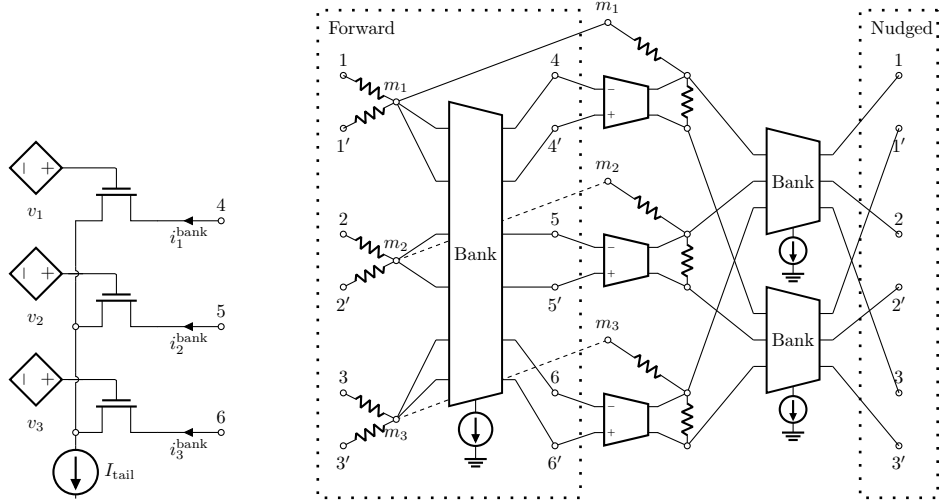

The reverse currents are more messy:
\begin{align*}
i_{a\to a'} & =\frac{I_{\mathrm{tail}}}{2\tau}\left[(z_{a+3}-z_{a+3}')-C\right]p_{a}\\
 & \quad C=(z_{4}-z'_{4})p_{1}+(z_{5}-z'_{5})p_{2}+(z_{6}-z'_{6})p_{3}.
\end{align*}
We tried a very crude approximation that ignores common channel $C$,
which can be implemented by a single VCCS per input $a$:
\[
i_{a\to a'}=\frac{I_{\mathrm{tail}}}{2\tau}(z_{a+3}-z_{a+3}').
\]
With this variant we found settings at which we could learn the ICL
task, but not the memorisation task of section \ref{sec:Training-for-memorisation}.
Thus at present we use a more complicated circuit shown in figure
\ref{fig:Bank-meangadget}. This is not ideal, it pulls current from
all $i_{a}$ and $i_{a'}$ for all inputs $(z_{a}-z_{a}')$, but the
TIA stage handles these currents well.

\subsection{Pair --- figure \ref{fig:Pair}}\label{subsec:Pair}

This is a variant of bank in which the tail current is one of the
inputs. Inputs are $v_{1},v_{2}=-v_{1},i_{3}$ and outputs are currents
at nodes 4,5:
\[
\begin{aligned}i_{4\to0} & =i_{3}\frac{e^{v_{1}/\tau}}{e^{v_{1}/\tau}+e^{v_{2}/\tau}}=i_{3}p_{1}\\
i_{5\to0} & =i_{3}\frac{e^{v_{2}/\tau}}{e^{v_{1}/\tau}+e^{v_{2}/\tau}}=i_{3}p_{2}.
\end{aligned}
\]
Here the block involves 3 currents
\begin{align}
f_{3}(x) & =i_{3}p_{1}+i_{3}p_{2}\nonumber \\
f_{4}(x) & =-i_{3}p_{1}\\
f_{5}(x) & =-i_{3}p_{2}\nonumber
\end{align}
anyway we arrive at dyadic potential
\[
h^{\mathrm{pair}}(z,z')=(z_{4}-z'_{4}-z_{3}+z'_{3})i_{3}p_{1}+(z_{5}-z'_{5}-z_{3}+z'_{3})i_{3}p_{2}
\]
and thus

\begin{align}
i_{1\to1'} & =(z_{4}-z'_{4}-z_{3}+z'_{3})\frac{i_{3}}{2\tau}p_{1}\nonumber \\
i_{2\to2'} & =(z_{5}-z'_{5}-z_{3}+z'_{3})\frac{i_{3}}{2\tau}p_{2}\nonumber \\
i_{3'\to3} & =\frac{i_{3}}{2V_{T}}\left[(z_{4}-z'_{4})p_{1}+(z_{5}-z'_{5})p_{2}\right]\\
 & =\frac{i_{3}}{2V_{T}}\left[\frac{z_{4}-z'_{4}+z_{5}-z'_{5}}{2}+\frac{z_{4}-z'_{4}-z_{5}+z'_{5}}{2}\right]\tanh\frac{z_{1}+z'_{1}-z_{2}-z'_{2}}{4\tau}\nonumber
\end{align}
We discard the $i_{1\to1'}$ terms (one layer) keep only the last
term, linearising the tanh around $v_{1}=v_{2}$.

\section{Cross-entropy loss as a floating circuit element}\label{sec:Cross-entropy-loss}

\begin{figure}
\centering \input{figures/mse.tikz}\hspace{1cm}\includegraphics{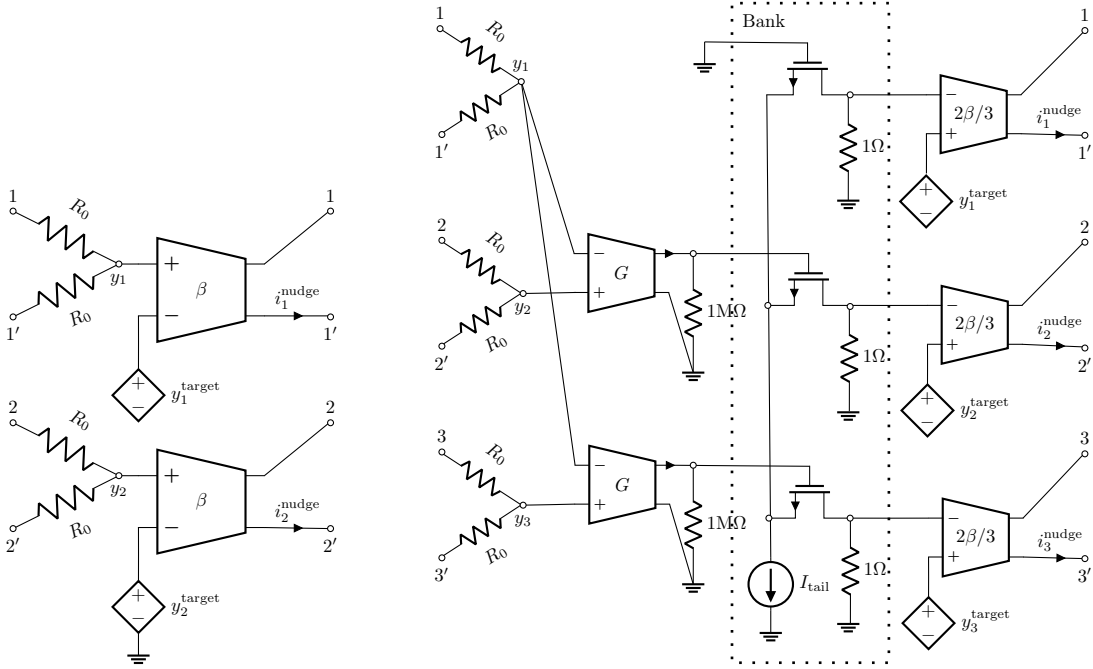}\caption{Loss-function elements. Left, generates the nudge appropriate for
MSE loss, as drawn in figure \ref{fig:Double-MLP} but now allowing
$m=2$ outputs. Each output node of the circuit is drawn twice: this
fragment reads the mean outputs $y_{1},y_{2}$, computes the nudge
current by comparing to the given $y^{\mathrm{target}}_{c}$, and
injects it back into the same nodes. Right, the equivalent for LCE
loss. The middle part is the ``bank'' circuit from figure \ref{fig:Bank};
its current outputs are converted to voltage by the $1\Omega$ resistors.
If driven directly by $y_{c}$ then the softmax temperature would
always be $nV_{T}=39$mV; the amplification stage lets us adjust this.
}\label{fig:Loss-MSE-LCE}
\end{figure}

Figure \ref{fig:Loss-MSE-LCE} left shows the same circuit element
for MSE as we had in figure \ref{fig:Double-MLP} in the main text,
now for two outputs $y_{1},y_{2}$. Note that we draw each output
node twice, so that the circuit reads left-to-right but influences
its input. Note also that the loss element for each $y_{c}$ is decoupled
from the others.

Figure \ref{fig:Loss-MSE-LCE} right shows a circuit implementing
the following nudge current, which is the gradient of LCE loss (\ref{eq:loss-LCE}):
\[
i^{\mathrm{nudge}}_{c\to c'}=\beta\:\frac{d}{dy_{c}}L_{\mathrm{LCE}}(y,y^{\mathrm{target}})=\frac{\beta}{\tau}\left[\softmax(y/\tau)_{c}-p^{\mathrm{target}}_{c}\right].
\]
The parts are as follows:
\begin{enumerate}
\item First, the mean output $y_{c}=(v_{c}+v_{c}')/2$ is found with a resistor
divider between $c$ and $c'$, with large fixed $R_{0}$.
\item These voltages are amplified, to allow control of the temperature
$\tau$. The first voltage, $y_{1}$, is used as the reference, and
replaced with 0. (Our circuit uses an OTA / VCCS device driving current
into a resistor to ground, thus essentially simulating an op-amp.)
\item These voltages are the input to the bank circuit from figure \ref{fig:Bank},
controlling the gates of $m$ transistors with a fixed total current
$I_{\mathrm{tail}}$. The resulting currents are $\softmax(y/\tau)_{c}$
.
\item The ``bank'' output currents each drain to ground, to produce voltages.
These voltages are compared to $p^{\mathrm{target}}_{c}$, supplied
as voltages from outside. The comparison involves a floating VCCS,
which drives current $i^{\mathrm{nudge}}_{c\to c'}$ between the same
output nodes.
\end{enumerate}

\section{Adaptive optimisation rules with memory}\label{sec:Adaptive-optimisation-rules}

Gradient descent is the most naiive way to use gradient information
to perform optimisation. Its two main limitations involve the shape
of the landscape, and how to deal with noise in the gradient vector.

The problem is roughly to find $\argmin_{\vec{\theta}}\mathcal{L}(\vec{\theta})$
where the objective is the mean of the loss function $L$ evaluated
for all points in the training dataset $\{(x_{i},y_{i}),i=1,2,\ldots N_{\mathrm{data}}\}$:
\[
\mathcal{L}(\vec{\theta})=\frac{1}{N_{\mathrm{data}}}\sum^{N_{\mathrm{data}}}_{i=1}L(f_{\vec{\theta}}(x_{i}),y_{i}).
\]
Here $f_{\vec{\theta}}(x)$ is a neural network, or the output of
a circuit with input voltages $x$. We write $\vec{\theta}\in\mathbb{R}^{M}$
to stress that this is all the parameters, for us a vector of all
the adjustable conductances $g_{ab}$. We write $\theta_{\mu}$ for
one element, with $\mu=1,2,\ldots M$.

In the neighbourhood of its minimum, $\mathcal{L}(\vec{\theta})$
might be some quadratic bowl, often greatly elongated \cite{mao2025analytical}.
If so, then knowing the Hessian at that point, $\partial^{2}\mathcal{L}(\vec{\theta})/\partial\theta_{\mu}\partial\theta_{\nu}$,
would allow descent to the minimum in one step. Computing all second
derivatives is very expensive, and methods like L-BFGS \cite{nocedal1980updating}
aim instead to estimate the Hessian from past gradient observations:
\[
\Delta_{\mu}(\vec{\theta})=\frac{d\mathcal{L}(\vec{\theta})}{d\theta_{\mu}},\qquad\text{ store }\{\Delta(\vec{\theta}_{t_{1}}),\Delta(\vec{\theta}_{t_{2}}),\ldots,\Delta(\vec{\theta}_{t_{\mathrm{limit}}})\}.
\]
One inspiration for methods like Adam is to estimate only the diagonal
elements of the Hessian. This involves $\mathcal{O}(M)$ numbers,
instead of $M^{2}$.

In machine learning, we typically do not find the gradient for all
training samples at once, but instead for just a small batch of them,
$1\leq N_{\mathrm{batch}}<N_{\mathrm{data}}$. And in the physical
training above, it's natural to use $N_{\mathrm{batch}}=1$:
\[
\delta_{\mu}(\vec{\theta})=\frac{dL(f_{\vec{\theta}}(x_{i}),y_{i})}{d\theta_{\mu}}
\]
This is a noisy measure of the true gradient $\Delta_{\mu}$, hence
this is described as stochastic gradient descent (SGD). Some noise
is thought to have beneficial properties, regularising the solution,
but not too much. One further function of optimisation rules is to
smooth this noise out, simulating a larger $N_{\mathrm{batch}}$ by
keeping a moving average of $\delta_{\mu}(\vec{\theta}_{t})$ for
some past times $t$, and updating based on this.

For a recent review of optimisation rules focused on this two-fold
role, preconditioning plus smoothing, see \cite{pethick2026training}.

\subsection{Adam \cite{kingma2017adam}}

For each scalar parameter $\theta\in\mathbb{R}$, Adam stores two
memories $m,v\in\mathbb{R}$, initially zero. These are estimates
of the first and second moments of the gradient $\bar{\theta}=dL/d\theta$,
updated as follows at each time-step $t=1,2,\ldots$, with decay rates
typically $\beta_{1}=0.9$ and $\beta_{2}=0.999$:
\[
\begin{aligned}m_{t+1} & =\beta_{1}m_{t}+(1-\beta_{1})\bar{\theta}_{t}\\
v_{t+1} & =\beta_{2}v_{t}+(1-\beta_{2})(\bar{\theta}_{t})^{2}.
\end{aligned}
\]
The parameter is updated as follows, with learning rate $\eta$:
\[
\theta_{t+1}=\theta_{t}-\eta\,\frac{m_{t+1}}{1-(\beta_{1})^{t}}\sqrt{\frac{1-(\beta_{2})^{t}}{v_{t+1}}}.
\]
The update is essentially $-\eta m_{t+1}/\sqrt{v_{t+1}}$; the other
factors are corrections for starting at $m_{0}=v_{0}=0$, which are
applied only for small $t$. Notice this invariance: a gradient (say)
10x larger makes $m$ 10x larger, and $v$ 100x larger, but has no
effect at all on the updates to $\theta$.

\subsection{Lion \cite{chen2023symbolic}}

This is a simpler optimiser, recently discovered by automated search.
For each scalar parameter $\theta\in\mathbb{R}$, it stores one memory
$\psi\in\mathbb{R}$. Then at each time-step $t$, given gradient
$\bar{\theta}=dL/d\theta$, the update rule is the following:
\[
\begin{aligned}\psi_{t+1} & =\beta_{2}\psi_{t}+(1-\beta_{2})\bar{\theta}_{t}\\
\theta_{t+1} & =\theta_{t}-\eta\sign\left[\beta_{1}\psi_{t}+(1-\beta_{1})\bar{\theta}_{t}\right].
\end{aligned}
\]
Here $\eta$ is the learning rate, and it sets the size of the fixed
steps $\theta_{t+1}=\theta_{t}\pm\eta$. By default $\beta_{2}=0.99$,
meaning that $\psi_{t}$ is a moving average of about 100 past gradients.
And $\beta_{1}=0.9$ means that a particular gradient which disagrees
with this average and is about 10 times larger will immediately move
the parameter. Clearly a gradient (say) 10x larger will, again, have
no effect on the updates to $\theta$.

If we set $\beta_{1}=\beta_{2}$ then Lion simplifies to use the memory
directly for the update:
\[
\begin{aligned}\psi_{t+1} & =\beta_{2}\psi_{t}+(1-\beta_{2})\bar{\theta}_{t}\\
\theta_{t+1} & =\theta_{t}-\eta\sign(\psi_{t+1}).
\end{aligned}
\]
This optimisation rule is what we're using for the physical training.
And this rule could plausibly be implemented as an RC circuit. A local
circuit for gradient descent must measure the voltage drops $\Delta v_{ab}$
and $\Delta v_{a'b'}$ (\ref{eq:gradient-double}) and use them to
update conductance $g_{ab}$ (\ref{eq:gradient-descent}). A local
circuit for Lion would instead have a capacitor which stores the memory
$\psi_{ab}$, with RC time about 100 training examples, loaded by
the gradient measurement. Then the conductance update has a fixed
rate of change controlled by the sign of the memory, the learning
rate, and the time $\Delta t$ between training examples:
\[
\begin{aligned}\frac{d\psi_{ab}}{dt} & =\frac{1-\beta_{2}}{\Delta t}\left[-\psi_{ab}+\frac{dL}{dg_{ab}}\right]\\
\frac{dg_{ab}}{dt} & =-\frac{\eta}{\Delta t}\:\sign(\psi_{ab}).
\end{aligned}
\]
We have not simulated a specific circuit for this, as we restrict
our simulations to DC. But it would not be much more complex to build
than the twin edge circuits of \cite{dillavou2024machine,handley2025microchips}.

\section{Transitions between ICL and memorisation}\label{sec:Training-for-memorisation}

\begin{figure}
\centering \includegraphics[width=14cm]{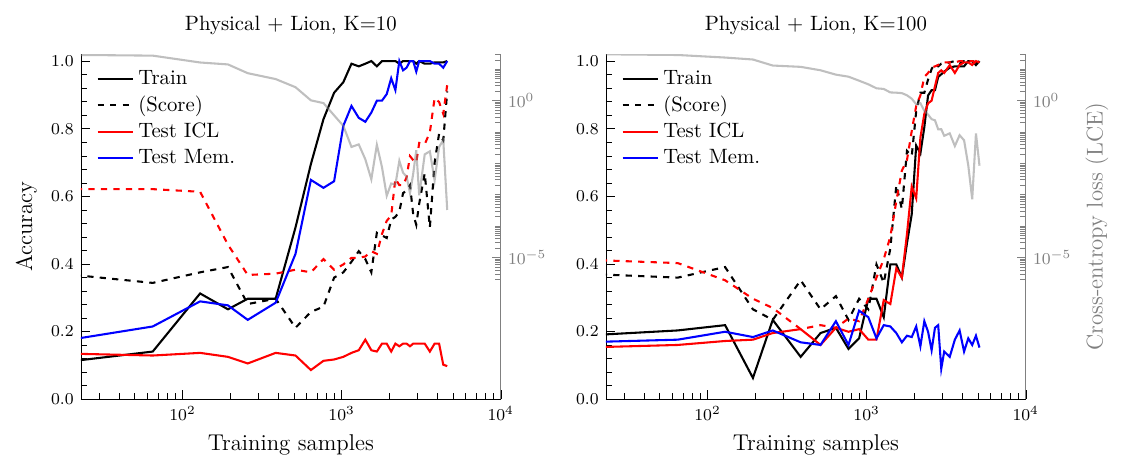}

\caption{Two training tasks: $K=10$ classes (``letters'') on the left, encouraging
memorisation, and $K=100$ on the right, identical to figure \ref{fig:Training-ICL}.
Here we use the physical approximate gradient elements. In addition
to accuracy on an ICL test (red, as before), we also plot a blue line
showing accuracy on a memorisation test (in which the answer does
not appear earlier in the string).}\label{fig:Train-K10-K100}
\end{figure}

The in-context learning (ICL) task of section \ref{sec:Minimal-ICL-task}
is adapted from Reddy \cite{reddy2024mechanistic,nguyen2025differential},
who showed a phase transition between generalisation and memorisation,
i.e. between ICL and ``in-weights learning''. This phase transition
also occurs in our transistor transformer, as we show in figure \ref{fig:Train-K10-K100}.
The rest of this appendix explains the task in more detail, and compares
to some other models.

Let us be more explicit about the setup. Training involves:
\begin{itemize}
\item A fixed set of class vectors, $\vec{x}^{(k)}\in\mathbb{R}^{d}$ for
$k=1,2,\ldots K$. These are the embedding vectors of the letters
in the main text. They are random, $\vec{x}^{(k)}\sim\mathcal{N}(0,1/d)$.
\item A fixed set of label vectors, $\vec{x}^{(\ell)}\in\mathbb{R}^{d'}$
for $\ell=1,2,\ldots L$. These are the embedding vectors of the numbers
in the main text. While \cite{reddy2024mechanistic} takes them to
be random, we take them to be ``one-hot'' vectors $\smash{x^{(\ell)}_{\mu}}=\delta_{\mu\ell}$.
Since dot-product attention (\ref{eq:attention}) is rotationally
symmetric, this involves little loss of generality, as random vectors
are nearly orthogonal for large $d'$. But our (\ref{eq:score-Bahdanau})
breaks this symmetry.
\item A fixed class-label relationship $k\mapsto\ell$, which is generally
many-to-one as $K>L$.
\item Each training sample is a sequence of $N$ class-label pairs, plus
one query class. In the training data, the query class always appears
earlier in the sequence. In \cite{reddy2024mechanistic} the sample
repeats $B$ times, described as a burst, but we take $B=1$. We take
$N=4$, and demand in addition that all $N$ pairs use distinct labels.
Thus random output has accuracy $1/L$, but simply returning any one
of the given sequence's labels gives accuracy $1/N$.
\item Noise is added to the class vectors, to ensure that for large enough
$K$ it is impossible to memorise the class-label relationship, as
different classes will overlap. As in \cite{reddy2024mechanistic}
we use $\vec{x}\sim[\vec{x}^{(k)}+\epsilon\mathcal{N}(0,1/d)]/\sqrt{1+\epsilon^{2}}$
and we take $\epsilon=0.1$.
\end{itemize}
There are two tests:
\begin{itemize}
\item The test for ICL involves a completely new set of class vectors, and
then constructs sequences as before.
\item The test for memorisation instead constructs sequences in which the
query class does not appear earlier in the sequence.
\end{itemize}
Test and training sequences are assembled into matrices as follows:
\begin{itemize}
\item For the two-layer model of \cite{reddy2024mechanistic}, the vectors
$\vec{x}^{(k_{1})},\vec{x}^{(\ell_{1})},\vec{x}^{(k_{2})},\vec{x}^{(\ell_{2})},\ldots$
are columns of a $d\times(2N+2)$ matrix (with $d=d'$). The full
$D\times(2N+2)$ input matrix concatenates a simple time-code, which
increments the position of one nonzero entry in each time-step, so
that $X_{\mu t}=1$ for $\mu=t+t_{0}\mod{D-d}$.
\item For the one-layer model of \cite{nguyen2025differential}, each pair
$\vec{x}^{(k_{t})},\vec{x}^{(\ell_{t})}$ is concatenated to one $D=d+d'$
vector, and these are columns of the full $D\times(N+1)$ input matrix.
There is no need for a time-code in this case.
\end{itemize}
The circuit described in section \ref{sec:Transformer-circuit} is
a one-layer model, and is capable of learning either strategy. Features
which increase the likelihood of learning ICL (because they make memorisation
harder) are increasing $K$, increasing burstiness $B$, and increasing
the noise.

Figure \ref{fig:Train-K10-K100} shows training with $K=10$ and $K=100$.
The case $K=100$ is identical to figure \ref{fig:Training-ICL} in
the main text, except that we plot the memorisation test accuracy
too.

For training accuracy and ICL test accuracy, solid lines use the output,
but dashed lines count whether the score $S_{sT}$ is highest at the
correct input time $s$, for $s<T$.

\subsection{Single circuits and the VF algorithm}

\begin{figure}
\centering \includegraphics[width=14cm]{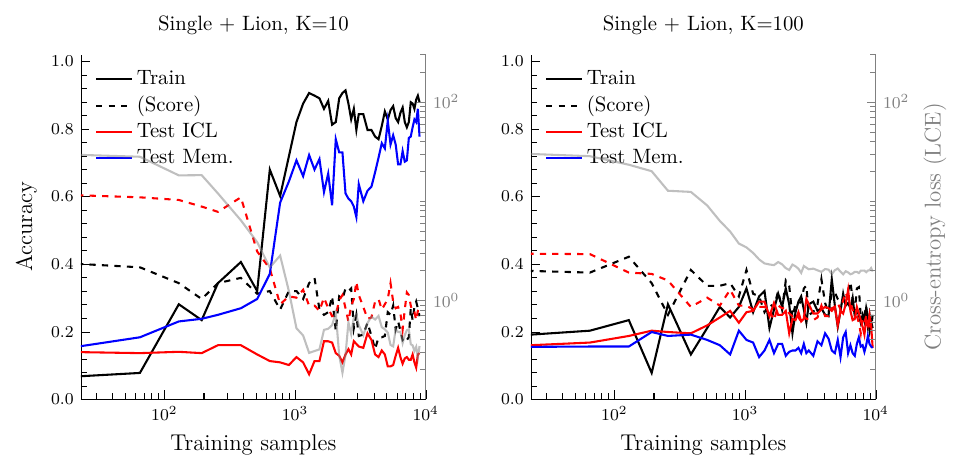}

\caption{Training using the VF algorithm, instead of the double circuit. At
left, $K=10$ encouraging memorisation, with partial success by training
the output value resistors alone. At right, $K=100$ should encourage
ICL, learns nothing, as no gradient information propagates to the
key and query resistors.}\label{fig:VF-alg}
\end{figure}

An early approach to applying equilibrium propagation to non-reciprocal
systems in \cite{scellier2018generalization} proposed simply using
(\ref{eq:gradient-nudge}) despite the lack of an energy function.
This is known as the vector field (VF) algorithm, and figure \ref{fig:VF-alg}
tries this with our transformer circuit, that is, applying a nudge
without doubling the circuit first. To first approximation this trains
only the output resistors, implementing $W^{\mathrm{value}}$. This
is sufficient for some memorisation to emerge, but the network cannot
learn to solve the ICL task without training the key and query resistors
which control attention.

\subsection{Memorisation in an MLP-like circuit}

\begin{figure}
\centering \includegraphics[width=14cm]{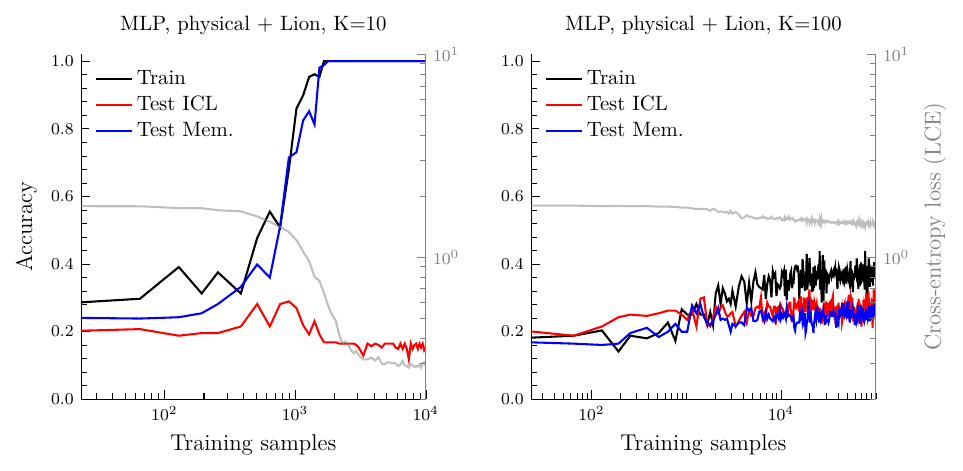}

\caption{Training a resistor MLP instead of the transistor transformer. At
left, $K=10$ encouraging memorisation. At right, $K=100$ should
encourage ICL, but it cannot learn this task. The circuit used is
as in \ref{fig:Single-MLP}, with 120 input nodes ($60=12\times5$
positive, and 60 inverted), 64 hidden nodes, 6 outputs. Even for $K=10^{4}$
and 512 hidden nodes we have not seen successful ICL.}\label{fig:MLP-no-ICL}
\end{figure}

Figure \ref{fig:MLP-no-ICL} shows the result of training a resistor
MLP on this task, precisely the model of \ref{fig:Single-MLP} except
enlarged to have $\vec{x}\in\mathbb{R}^{5\times12}$ i.e. 120 input
nodes, 64 hidden nodes, and $\vec{y}\in\mathbb{R}^{6}$ as 6 output
nodes. This has 4352 trainable resistors, an order of magnitude larger
than our transformer. It has no difficulty memorising $K=10$ letters,
but we have not seen it learn to perform ICL.

\subsection{Convergent Energy Transformer \cite{hoier2026training}}\label{subsec:CET}

We adapted the CET model from \cite{hoier2026training} to suit our
task, with matrix input $X_{\mu t}$ ($t=1,2,\ldots,T$, $\mu,\nu=1,2,\ldots,D$)
and vector output $y_{\delta}$ ($\delta=1,2,\ldots,D_{\mathrm{out}}$).
Their energy function can be written
\[
\begin{aligned}E_{\theta}(X,y,Z)= & \sum_{\mu,t}\left[\frac{1}{2}Z^{\,2}_{\mu t}-Z_{\mu t}\left(\smash{\sum_{\nu}}W^{\mathrm{encoder}}_{\mu\nu}X_{\nu t}+b^{\mathrm{encoder}}_{\mu}+b^{\mathrm{position}}_{\mu t}\right)\right]\\[1mm]
 & -\frac{1}{\gamma}\sum_{t}\log\sum_{s}\exp\gamma\sum_{\mu\nu\alpha}(W^{\mathrm{query}}_{\alpha\mu}Z_{\mu t})(W^{\mathrm{key}}_{\alpha\nu}Z_{\nu s})\\[1mm]
 & -\sum_{\kappa,t}\relu\left(\smash{\sum_{\mu}}W^{\mathrm{memory}}_{\mu\kappa}Z_{\mu t}\right)^{2}\\[1mm]
 & +\sum_{\delta}\left[\frac{1}{2}y^{\,2}_{\delta}-y_{\delta}\left(\smash{\sum_{\nu}}W^{\mathrm{decoder}}_{\mu\delta}Z_{\mu T}+b^{\mathrm{decoder}}_{\delta}\right)\right].
\end{aligned}
\]
The parameters are
\[
\theta=\left\{ W^{\mathrm{encoder}}_{\mu\nu},b^{\mathrm{encoder}}_{\mu},b^{\mathrm{position}}_{\mu t},W^{\mathrm{query}}_{\alpha\mu},W^{\mathrm{key}}_{\alpha\mu},W^{\mathrm{memory}}_{\mu\kappa},W^{\mathrm{decoder}}_{\mu\delta},b^{\mathrm{decoder}}_{\delta}\right\}
\]
and $\alpha=1,2,\ldots,D_{\mathrm{head}}=D$ (single-headed) and $\kappa=1,2,\ldots,D_{\mathrm{mem}}$
(Hopfield memories). The forward evaluation of the model is then
\[
f(X)=\argmin_{y}\min_{Z}E_{\theta}(X,y,Z)
\]
with the hidden state constrained by
\[
\sum_{\mu}Z_{\mu t}=0,\qquad\frac{1}{D}\sum_{\mu}Z^{\,2}_{\mu t}=1.
\]

Our data has $D=12$, $D_{\mathrm{out}}=6$ and $T=5$ as before,
and we vary $D_{\mathrm{mem}}\in\{24,48,96\}$ and $\gamma\in\{1/16,1/4,1\}$.
We train using cross-entropy loss, exact gradients, and Adam. We see
perfect memorisation of the data with $K=10$ classes, and ICL accuracy
at chance for $K=100$. That is, it behaves like the MLP of figure
\ref{fig:MLP-no-ICL}.

\subsection{Artificial neural networks}\label{subsec:Artificial-neural-networks}

\begin{figure}
\centering \includegraphics[width=15cm]{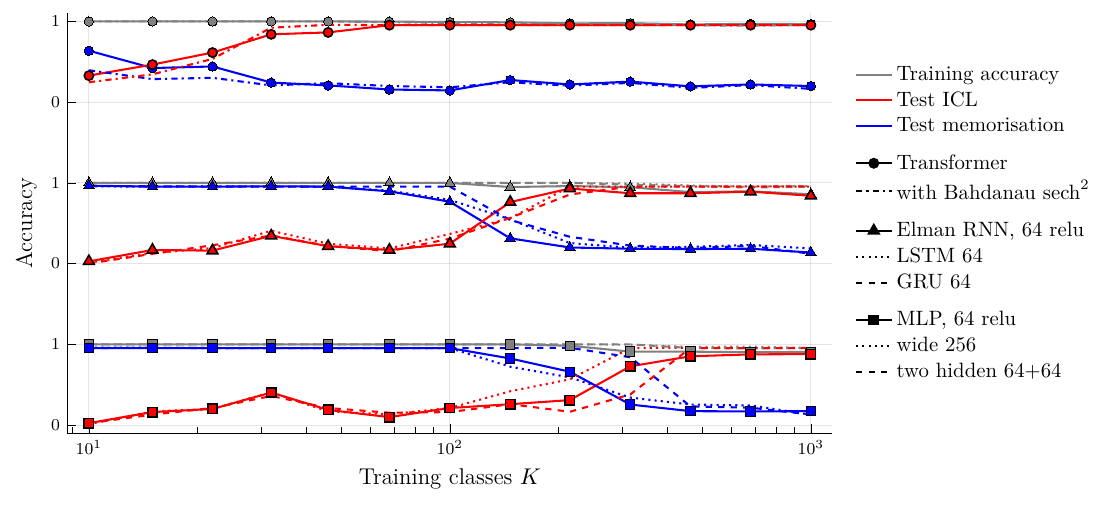}

\caption{Transitions between memorisation and ICL for several artificial neural
networks, using the same training task. Transformer variants (top)
use (\ref{eq:attention}) and (\ref{eq:score-Bahdanau}), MLP variants
(bottom) are (\ref{eq:MLP}) with 64 and 256 hidden nodes, and one
with a second hidden layer. Three standard kinds of recurrent neural
networks \cite{elman1990finding,hochreiter1997long,cho2014learning}
are also shown (middle). Note that the models have widely varying
parameter counts: 432 and 300 for the transformers, 4--$8\times10^{3}$
for the MLPs, 5--$20\times10^{3}$ for the RNNs.  }\label{fig:Small-sweep}
\end{figure}

Our statement above that MLPs struggle to learn this task ought to
be quantified somehow. While ICL is often described as a key characteristic
of how transformers work, it is not impossible to train an MLP to
perform ICL, but it does require encouragement. While results above
use $K=10$ to see memorisation and $K=100$ to produce ICL, figure
\ref{fig:Small-sweep} explores much larger $K$. Here we use ordinary
digital neural networks. Similar results were obtained by \cite{tong2025mlps}.

Figure \ref{fig:Small-sweep} also includes several kinds of recurrent
neural networks (RNNs), and sees intermediate performance for these.
The literature has varied results on RNNs learning ICL tasks, see
for instance \cite{chan2022datadistributional,arora2023zoology,lee2024attention}.
There are also results on convolutional networks such as \cite{kantroo2025incontext}.

\section{Nonlinear circuit solver}\label{sec:Nonlinear-circuit-solvers}

Finding the operating point of linear circuits amounts to one matrix
inversion. For the piecewise linear circuits like figure \ref{fig:Single-MLP},
we can use the fast algorithm of \cite{scellier2024fast} --- with
the caveat that it only handles diodes connected to a fixed voltage.

For the truly nonlinear circuits, we use a custom Newton-Raphson solver.
The MOSFET circuits are badly conditioned, as the exponentials can
take a wide range of values. Tracking a tiny nudge on top of that,
accurately, is a challenge. The solution used in our solver is to
change variables, and re-write the circuit equations in terms of $(v_{a}+v_{a}')/2$
and $v_{a}-v_{a}'$. There is no physical change here, and we do not
for instance demand that $(v_{a}+v_{a}')/2$ is unchanged by the nudge.

The most common off-the-shelf solver is SPICE and its descendants
\cite{nagel1973spice,vogt2026ngspice}. We have built a bridge across
to this, and have confirmed that Ngspice is able to solve the final
model produced by training (figure \ref{fig:Training-ICL} ``physical'')
and achieves the same accuracy with the same ICL test data. We still
use ideal components, especially transistors obeying (\ref{eq:current-mosfet})
instead of a fully realistic law.

\end{document}

%% file: figures/denver.tikz
\begin{tikzpicture}[transform shape, scale=0.7]
	\draw (15, -10) to[american controlled voltage source, l={$v_1=x_1$}] (14, -10);
	\draw (14, -12) to[american controlled voltage source, l_={$v_2=-x_1$}] (15, -12);
	\draw (15, -14) to[american controlled voltage source, l={$v_3=x_2$}] (14, -14);
	\draw (15, -10) to[american resistor, l={$r_{14}$}] (17, -11);
	\draw (15, -12) to[american resistor, l_={$r_{24}$}] (17, -11);
	\draw (17, -11) to[short] (18.246, -12.24);
	\draw (15, -14) to[short] (18.013, -12.607);
	\draw (19.25, -12.25) to[short, i^>=${i = K v_3 v_4}$] (21.75, -12.25);
	\draw (21.75, -12.25) to[american resistor, l={$r_{05}$}] (21.75, -14.5);
	\node[ground] at (21.75, -14.25){};
	\draw (19.5, -12.75) -- (20, -12.75) -| (20, -14.25);
	\node[ground] at (20, -14.25){};
	\draw (18, -12.25) -- (17, -12.25) -| (17, -14.25);
	\draw (18.33, -13.019) -| (17.75, -13.25) -| (17.75, -14.25);
	\node[ground] at (17.75, -14.25){};
	\node[ground] at (17, -14.25){};
	\node[ocirc](N1) at (17, -11){} node[anchor=south] at (N1.north){$4$};
	\draw[fill={rgb,255:red,255;green,255;blue,255}, line width=1pt] (17.75, -12.5) -- (18, -11.75) -- (19.5, -12) -- (19.5, -13) -- (18, -13.25) -- cycle node[anchor=center] at (18.625, -12.5){$\ \quad K$};
	\node[ocirc](N2) at (15, -14){} node[anchor=south] at (N2.north){$3$};
	\node[ocirc](N3) at (15, -12){} node[anchor=south] at (N3.north){$2$};
	\node[ocirc](N4) at (15, -10){} node[anchor=south] at (N4.north){$1$};
	\node[ocirc](N5) at (21.75, -12.25){} node[anchor=south] at (N5.north){$5$};
	\node[ground] at (13.5, -14.25){};
	\draw (13.5, -14.25) to[short] (13.5, -10);
	\draw (14, -10) to[short] (13.5, -10);
	\draw (14, -12) to[short] (13.5, -12);
	\draw (14, -14) to[short] (13.5, -14);
	\node[shape=rectangle, minimum width=1.965cm, minimum height=1.215cm] at (21.5, -10.625){} node[anchor=north, align=center, text width=1.577cm, inner sep=6pt] at (21.5, -10){Output: $y=v_5$};
\end{tikzpicture}

%% file: figures/denver_double_red.tikz
\begin{tikzpicture}[transform shape, scale=0.7]
	\draw (19.5, -16) to[short] (18.75, -16);
	\draw (20.75, -13.75) to[short] (20, -13.5);
	\draw (14.5, -10) to[american controlled voltage source, l={$v_1=x_1$}] (13.5, -10);
	\draw (13.5, -12) to[american controlled voltage source, l_={$v_2=-x_1$}] (14.5, -12);
	\draw (14.5, -14) to[american controlled voltage source, l={$v_3=x_2$}] (13.5, -14);
	\draw (14.5, -10) to[american resistor, l={${r_{14} = r_{1'4'}}$}] (17, -10);
	\draw (14.5, -12) to[american resistor] (17, -10);
	\draw (17.004, -10.01) to[short] (18.875, -9.125);
	\draw (14.5, -14) to[short] (17, -14);
	\draw (20.5, -9.5) to[short, i^>=${\ }$, current/distance=0.2] (23, -11);
	\draw (23, -13) to[american resistor] (23.75, -15.5);
	\node[ground] at (23.75, -15.5){};
	\draw[fill={rgb,255:red,255;green,255;blue,255}, line width=0.8pt] (18.75, -9.5) -- (19, -8.75) -- (20.5, -9) -- (20.5, -10) -- (19, -10.25) -- cycle node[anchor=center] at (19.625, -9.5){$\ \quad K$};
	\draw (14.5, -10) to[american resistor] (17, -12);
	\draw (14.5, -12) to[american resistor, l_={${r_{2'4'}=r_{24}}$}] (17, -12);
	\draw[fill={rgb,255:red,255;green,255;blue,255}, line width=0.8pt] (18.75, -11.5) -- (19, -10.75) -- (20.5, -11) -- (20.5, -12) -- (19, -12.25) -- cycle node[anchor=center] at (19.625, -11.5){$\ \quad K$};
	\draw (17, -12) to[short] (18.875, -11.125);
	\draw (20.5, -11.5) to[short, i^>=${\ }$, current/distance=0.2] (23, -13);
	\draw (17, -14) to[short] (18.875, -9.875);
	\draw (17, -14) to[short] (18.875, -11.875);
	\draw (17.004, -10.01) to[short, i^<=${\ }$, current/distance=0.85] (18.75, -13.25);
	\draw (17, -12) to[short] (18.75, -13.75);
	\draw (20, -13.25) to[short] (23, -11);
	\draw (20, -13.25) to[short] (23, -13);
	\draw (17, -10) to[short] (19, -15);
	\draw (17, -12) to[short] (19, -15.5);
	\draw (17, -14) to[short] (19, -16);
	\draw (20.5, -15.25) to[short, i^>=${\ }$, current/distance=0.2] (23, -11);
	\draw (20.5, -15.75) to[short] (23, -13);
	\draw (23, -11) to[american resistor, l={${r_{05}=r_{05'}}$}] (24.5, -15.5);
	\node[ground] at (24.5, -15.5){};
	\draw[fill={rgb,255:red,255;green,255;blue,255}, line width=0.8pt] (18.75, -15.5) -- (19, -14.75) -- (20.5, -15) -- (20.5, -16) -- (19, -16.25) -- cycle node[anchor=center] at (19.625, -15.5){$\ \quad K/2$};
	\node[shape=rectangle, minimum width=2.465cm, minimum height=1.965cm] at (24.25, -9.75){} node[anchor=north, align=center, text width=2.077cm, inner sep=6pt] at (24.25, -8.75){Output: $y = \dfrac{v_5 + v_5'}{2}$};
	\draw (17, -14) to[short] (20.5, -14.5);
	\draw (20.375, -13.875) to[short] (20.5, -14.5);
	\node[ocirc](N1) at (14.5, -12){} node[anchor=south] at (N1.north){$2$};
	\node[ocirc](N2) at (14.5, -10){} node[anchor=south] at (N2.north){$1$};
	\node[ground] at (18.75, -16){};
	\node[ground] at (20.75, -13.75){};
	\path[draw={rgb,255:red,255;green,38;blue,0}, line width=0.5pt] (23, -11) -- (20, -13.25) -- (23, -13);
	\path[draw={rgb,255:red,255;green,38;blue,0}, line width=0.5pt] (20.375, -13.875) -- (20.5, -14.5) -- (17, -14);
	\node[ocirc](N3) at (23, -11){} node[anchor=south] at (N3.north){$5$};
	\node[ocirc](N4) at (23, -13){} node[anchor=south] at (N4.north){$5'$};
	\node[ocirc](N5) at (17, -14){} node[anchor=south east] at (N5.north west){$3 = 3'$};
	\path[draw={rgb,255:red,255;green,38;blue,0}, line width=0.5pt] (18.75, -13.25) -- (17, -10);
	\node[ocirc](N6) at (17, -10){} node[anchor=south] at (N6.north){$4$};
	\path[draw={rgb,255:red,255;green,38;blue,0}, line width=0.5pt] (20.75, -13.75) -- (20, -13.5);
	\path[draw={rgb,255:red,255;green,38;blue,0}, line width=0.5pt] (18.75, -13.75) -- (17, -12);
	\node[ocirc](N7) at (17, -12){} node[anchor=south] at (N7.north){$4'$};
	\path[fill={rgb,255:red,255;green,255;blue,255}, draw={rgb,255:red,255;green,38;blue,0}, line width=0.8pt] (20.5, -13.5) -- (20.25, -14.25) -- (18.75, -14) -- (18.75, -13) -- (20.25, -12.75) -- cycle node[anchor=center] at (19.625, -13.5){\textcolor{rgb,255:red,255;green,38;blue,0}{${K/2 \quad \ }$}};
\end{tikzpicture}

%% file: figures/bump.tikz
\begin{tikzpicture}[transform shape, scale=0.7]
	\draw (4, 7) to[american controlled voltage source, l={$v_1$}] (3, 7);
	\draw (4, 4.5) to[american controlled voltage source, l={$v_2$}] (3, 4.5);
	\node[nmos, rotate=-90] at (5.02, 3.5){};
	\node[nmos, rotate=-90] at (5.02, 6){};
	\draw (6.5, 6) to[short, i^>=${\ }$] (5.79, 6);
	\node[jump crossing, rotate=-90] at (4.25, 4.5){};
	\draw (4, 4.5) to[short] (9, 4.5);
	\draw (4, 7) to[short] (7.5, 7);
	\draw (4.25, 4.36) to[short] (4.25, 3);
	\draw (4.25, 6) to[short] (4.25, 4.64);
	\draw (11.04, 5.5) to[short, i^>=$i^\mathrm{bump}$] (9.83, 5.5);
	\node[ocirc] at (4.25, 3.5){};
	\draw (4.25, 3) to[american current source, l={$I_\mathrm{tail}$}] (4.25, 2);
	\draw (5, 7) to[short] (5.02, 6.98);
	\draw (5, 4.5) to[short] (5.02, 4.48);
	\node[ground] at (4.25, 2){};
	\node[nmos, rotate=-90] at (7.52, 5.5){};
	\node[nmos, rotate=90, yscale=-1] at (9.06, 5.5){};
	\draw (4.25, 5.5) to[short] (6.75, 5.5);
	\draw (6.5, 3.5) to[short, i^>=${\ }$] (5.79, 3.5);
	\draw (9, 4.5) to[short] (9.06, 4.52);
	\draw (7.5, 7) to[short] (7.52, 6.48);
	\node[ground] at (6.5, 3.5){};
	\node[ocirc] at (4.25, 5.5){};
	\node[jump crossing, rotate=-90] at (6.5, 5.5){};
	\node[ground] at (6.5, 5.36){};
	\draw (6.5, 6) to[short] (6.5, 5.64);
	\node[ocirc](N1) at (11, 5.5){} node[anchor=south] at (N1.north){$5$};
	\node[ocirc](N2) at (6.5, 6){} node[anchor=south] at (N2.north){$3$};
	\node[ocirc](N3) at (6.5, 3.5){} node[anchor=south] at (N3.north){$4$};
	\draw (8.71, 5.5) to[short, i^>=${\ }$] (8, 5.5);
\end{tikzpicture}

%% file: figures/bank.tikz
\begin{tikzpicture}[transform shape, scale=0.7]
	\draw (3, 3) to[short] (4, 3);
	\draw (3, 5) to[short] (4, 5);
	\draw (3, 7) to[american controlled voltage source, l={$v_1$}] (2, 7);
	\draw (3, 5) to[american controlled voltage source, l={$v_2$}] (2, 5);
	\draw (3, 3) to[american controlled voltage source, l={$v_3$}] (2, 3);
	\node[nmos, rotate=-90] at (4.02, 4){};
	\node[nmos, rotate=-90] at (4.02, 6){};
	\node[nmos, rotate=-90] at (4.02, 2){};
	\draw (6, 6) to[short, i^>=$i_1^\mathrm{bank}$] (4.79, 6);
	\node[jump crossing, rotate=-90] at (3.25, 3){};
	\node[jump crossing, rotate=-90] at (3.25, 5){};
	\draw (3, 7) to[short] (4, 7);
	\draw (3.25, 4.86) to[short] (3.25, 3.14);
	\draw (3.25, 2.86) to[short] (3.25, 1.75);
	\draw (3.25, 6) to[short] (3.25, 5.14);
	\draw (6, 4) to[short, i^>=$i_2^\mathrm{bank}$] (4.79, 4);
	\draw (6, 2) to[short, i^>=$i_3^\mathrm{bank}$] (4.79, 2);
	\node[ocirc] at (3.25, 4){};
	\node[ocirc] at (3.25, 2){};
	\draw (3.25, 1.75) to[american current source, l={$I_\mathrm{tail}$}] (3.25, 0.75);
	\draw (4, 7) to[short] (4.02, 6.98);
	\draw (4, 5) to[short] (4.02, 4.98);
	\draw (4.02, 2.98) to[short] (4, 3);
	\node[ocirc](N1) at (6, 6){} node[anchor=south] at (N1.north){$4$};
	\node[ocirc](N2) at (6, 4){} node[anchor=south] at (N2.north){$5$};
	\node[ocirc](N3) at (6, 2){} node[anchor=south] at (N3.north){$6$};
\end{tikzpicture}

%% file: figures/timp.tikz
\begin{tikzpicture}[transform shape, scale=0.7]
	\draw (28.25, -15) to[american resistor, l_={$R$}] (24, -15);
	\node[fd inst amp, xscale=0.8, yscale=0.8](vccs) at (25.202, -16.392){} node[anchor=center] at (vccs.text){$G$};
	\draw (28.25, -15) to[short] (28.25, -16);
	\draw (23.5, -15) to[american controlled current source, l={${\text{Input }i_1}$}] (22.5, -15);
	\draw (26.154, -16) to[short, i^>=${i=-Gv_1}$, current/distance=0.4] (28.25, -16);
	\draw (26.154, -16.784) to[short] (26.404, -16.784);
	\draw (26.404, -16.784) to[short] (26.404, -17.284);
	\draw (23.5, -15) to[short] (24, -15);
	\draw (24, -15) to[short] (24, -16);
	\node[ocirc](N1) at (24, -15){} node[anchor=south] at (N1.north){$1$};
	\node[tlground] at (26.404, -17.284){};
	\node[tlground] at (28.25, -17.25){};
	\draw (28.25, -16) to[qvprobe, l={${\text{Output }v_2}$}] (28.25, -17.25);
	\node[ground] at (24, -16.784){};
	\draw (24.25, -16.784) to[short] (24, -16.784);
	\node[ocirc](N2) at (28.25, -16){} node[anchor=south west] at (N2.north east){$2$};
	\draw (24, -16) to[short] (24.25, -16);
\end{tikzpicture}

%% file: figures/pair.tikz
\begin{tikzpicture}[transform shape, scale=0.7]
	\draw (4, 8) to[american controlled voltage source, l={${v_1=X_{\mu s}}$}] (3, 8);
	\draw (4, 4) to[american controlled current source, mirror, l={${i_s^\mathrm{bank}}$}] (3, 4);
	\draw (4, 6) to[american controlled voltage source, l={${v_2=-X_{\mu s}}$}] (3, 6);
	\node[nmos, rotate=-90] at (5.77, 7){};
	\node[nmos, rotate=-90] at (5.77, 5){};
	\draw (5, 4) to[short] (5, 5.86);
	\draw (5, 6.14) to[short] (5, 7);
	\draw (4, 6) to[short] (5.75, 6);
	\draw (4, 8) to[short] (5.75, 8);
	\draw (4, 4) to[short] (5, 4);
	\draw (6.54, 7) to[short, i_<=${i_+^\mathrm{pair}}$] (8.04, 7);
	\draw (6.54, 5) to[short, i_<=${i_-^\mathrm{pair}}$] (8.04, 5);
	\node[jump crossing, rotate=-90] at (5, 6){};
	\draw (5.75, 8) to[short] (5.77, 7.98);
	\draw (5.75, 6) to[short] (5.77, 5.98);
	\node[ocirc](N1) at (5, 5){} node[anchor=east] at (N1.west){$3$};
	\node[ocirc](N2) at (8, 7){} node[anchor=south] at (N2.north){$4$};
	\node[ocirc](N3) at (8.04, 5){} node[anchor=south] at (N3.north){$5$};
\end{tikzpicture}

%% file: figures/bank_meangadget.tikz
\begin{tikzpicture}[transform shape, scale=0.7]
	\node[shape=rectangle, draw, line width=1pt, dash pattern={on 1pt off 4pt}, minimum width=1.465cm, minimum height=9.215cm] at (11.5, 4.625){} node[anchor=north west, align=left, text width=1.077cm, inner sep=6pt] at (10.75, 9.25){Nudged};
	\node[shape=rectangle, draw, line width=1pt, dash pattern={on 1pt off 4pt}, minimum width=4.965cm, minimum height=9.215cm] at (3, 4.625){} node[anchor=north west, align=left, text width=4.577cm, inner sep=6pt] at (0.5, 9.25){Forward};
	\draw (2.75, 6) to[short] (4.25, 6);
	\draw (2.75, 7) to[short] (4.25, 7);
	\draw[dash pattern={on 1.6pt off 1.6pt}] (2, 1.5) -- (6, 3);
	\draw[dash pattern={on 1.6pt off 1.6pt}] (2, 4.5) -- (6, 6);
	\draw (5, 8) to[short] (5.75, 7.74);
	\draw (5, 7.01) to[short] (5.75, 7.25);
	\draw (5, 5.01) to[short] (5.75, 4.75);
	\draw (5, 4) to[short] (5.75, 4.26);
	\draw (5, 2) to[short] (5.75, 1.74);
	\draw (5, 1) to[short] (5.75, 1.25);
	\draw (2, 4.5) to[short] (2.75, 4);
	\draw (2, 4.5) to[short] (2.75, 5);
	\draw (2, 7.5) to[short] (2.75, 7);
	\draw (2, 7.5) to[short] (6, 9);
	\draw (2, 7.5) to[short] (2.75, 6);
	\draw (2, 1.5) to[short] (2.75, 3);
	\draw (2, 1.5) to[short] (2.75, 2.25);
	\draw (4.25, 7) to[short] (5, 8);
	\draw (4.25, 6) to[short] (5, 7);
	\draw (2.75, 5) to[short] (5, 5);
	\draw (2.75, 4) to[short] (5, 4);
	\draw (4.25, 3) to[short] (5, 2);
	\draw (4.25, 2.25) to[short] (5, 1);
	\draw (1, 2) to[american resistor, /tikz/circuitikz/bipoles/length=0.700cm] (2, 1.5);
	\draw (1, 1) to[american resistor, /tikz/circuitikz/bipoles/length=0.700cm] (2, 1.5);
	\draw (1, 5) to[american resistor, /tikz/circuitikz/bipoles/length=0.700cm] (2, 4.5);
	\draw (1, 4) to[american resistor, /tikz/circuitikz/bipoles/length=0.700cm] (2, 4.5);
	\node[ocirc](N1) at (1, 8){} node[anchor=south] at (N1.north){$1$};
	\node[ocirc](N2) at (1, 7){} node[anchor=north] at (N2.south){$1'$};
	\node[ocirc](N3) at (1, 5){} node[anchor=south] at (N3.north){$2$};
	\node[ocirc](N4) at (1, 4){} node[anchor=north] at (N4.south){$2'$};
	\node[ocirc](N5) at (1, 2){} node[anchor=south] at (N5.north){$3$};
	\node[ocirc](N6) at (1, 1){} node[anchor=north] at (N6.south){$3'$};
	\node[ocirc](N7) at (2, 4.5){} node[anchor=south] at (N7.north){$m_2$};
	\node[ocirc](N8) at (5, 8){} node[anchor=south] at (N8.north){$4$};
	\node[ocirc](N9) at (5, 7){} node[anchor=north] at (N9.south){$4'$};
	\node[ocirc](N10) at (5, 5){} node[anchor=south] at (N10.north){$5$};
	\node[ocirc](N11) at (5, 4){} node[anchor=north] at (N11.south){$5'$};
	\node[ocirc](N12) at (5, 2){} node[anchor=south] at (N12.north){$6$};
	\node[ocirc](N13) at (5, 1){} node[anchor=north] at (N13.south){$6'$};
	\node[fd inst amp, xscale=0.5, yscale=0.5] at (6.345, 1.495){};
	\node[fd inst amp, xscale=0.5, yscale=0.5] at (6.345, 4.505){};
	\draw (1, 8) to[american resistor, /tikz/circuitikz/bipoles/length=0.700cm] (2, 7.5);
	\draw (1, 7) to[american resistor, /tikz/circuitikz/bipoles/length=0.700cm] (2, 7.5);
	\node[fd inst amp, xscale=0.5, yscale=0.5] at (6.345, 7.495){};
	\draw (6, 9) to[american resistor, /tikz/circuitikz/bipoles/length=0.700cm] (7.5, 8);
	\node[ocirc](N14) at (11.5, 8){} node[anchor=south] at (N14.north){$1$};
	\node[ocirc](N15) at (11.5, 7){} node[anchor=north] at (N15.south){$1'$};
	\node[ocirc](N16) at (11.5, 5){} node[anchor=south] at (N16.north){$2$};
	\node[ocirc](N17) at (11.5, 4){} node[anchor=north] at (N17.south){$2'$};
	\node[ocirc](N18) at (11.5, 2){} node[anchor=south] at (N18.north){$3$};
	\node[ocirc](N19) at (11.5, 1){} node[anchor=north] at (N19.south){$3'$};
	\draw (7.5, 8) to[american resistor, /tikz/circuitikz/bipoles/length=0.700cm] (7.5, 7);
	\draw (10.25, 6.5) to[short] (11.5, 8);
	\draw (10.25, 5.5) to[short] (11.5, 2);
	\draw (7.5, 8) to[short] (8.75, 6.5);
	\draw (7.5, 5) to[short] (8.75, 6);
	\draw (7.5, 2) to[short] (8.75, 5.5);
	\draw (6, 6) to[american resistor, /tikz/circuitikz/bipoles/length=0.700cm] (7.5, 5);
	\draw (6, 3) to[american resistor, /tikz/circuitikz/bipoles/length=0.700cm] (7.5, 2);
	\draw (7.5, 5) to[american resistor, /tikz/circuitikz/bipoles/length=0.700cm] (7.5, 4);
	\draw (7.5, 2) to[american resistor, /tikz/circuitikz/bipoles/length=0.700cm] (7.5, 1);
	\draw (7.5, 1) to[short] (8.75, 2.5);
	\draw (7.5, 7) to[short] (8.75, 3.5);
	\node[ocirc](N20) at (2, 7.5){} node[anchor=south] at (N20.north){$m_1$};
	\node[ocirc](N21) at (6, 9){} node[anchor=south] at (N21.north){$m_1$};
	\node[ocirc](N22) at (6, 6){} node[anchor=south] at (N22.north){$m_2$};
	\node[ocirc](N23) at (6, 3){} node[anchor=south] at (N23.north){$m_3$};
	\draw (6.94, 7.25) to[short] (7.5, 7);
	\draw (6.94, 7.74) to[short] (7.5, 8);
	\draw (10.25, 6) to[short] (11.5, 5);
	\draw (10.25, 3.5) to[short] (11.5, 7);
	\draw (10.25, 3) to[short] (11.5, 4);
	\draw (10.25, 2.5) to[short] (11.5, 1);
	\draw (7.5, 4) to[short] (8.75, 3);
	\draw (3.5, 1.875) to[american current source, /tikz/circuitikz/bipoles/length=1.05cm] (3.5, 0.75);
	\node[tlground] at (3.5, 0.75){};
	\draw (9.5, 5.125) to[american current source, /tikz/circuitikz/bipoles/length=0.840cm] (9.5, 4.25);
	\node[tlground] at (9.5, 4.25){};
	\draw (9.5, 2.125) to[american current source, /tikz/circuitikz/bipoles/length=0.840cm] (9.5, 1.25);
	\node[tlground] at (9.5, 1.25){};
	\draw (6.94, 4.75) to[short] (7.5, 5);
	\draw (6.94, 4.26) to[short] (7.5, 4);
	\draw (6.94, 1.74) to[short] (7.5, 2);
	\draw (6.94, 1.25) to[short] (7.5, 1);
	\node[ocirc] at (7.5, 8){};
	\node[ocirc] at (7.5, 7){};
	\node[ocirc] at (7.5, 5){};
	\node[ocirc] at (7.5, 4){};
	\node[ocirc] at (7.5, 2){};
	\node[ocirc] at (7.5, 1){};
	\draw (2.75, 3) to[short] (4.25, 3);
	\draw (2.75, 2.25) to[short] (4.25, 2.25);
	\node[ocirc](N24) at (2, 1.5){} node[anchor=north] at (N24.south){$m_3$};
	\draw[fill={rgb,255:red,255;green,255;blue,255}, line width=0.75pt] (3, 7.5) -- (3, 1.75) -- (4, 2) -- (4, 7.25) -- cycle node[anchor=center] at (3.5, 4.625){$\text{Bank}$};
	\draw (8.75, 6.5) to[short] (10.25, 6.5);
	\draw (8.75, 6) to[short] (10.25, 6);
	\draw (8.75, 5.5) to[short] (10.25, 5.5);
	\draw[fill={rgb,255:red,255;green,255;blue,255}, line width=0.75pt] (9, 7) -- (9, 5) -- (10, 5.25) -- (10, 6.75) -- cycle node[anchor=center] at (9.5, 6){$\text{Bank}$};
	\draw (8.75, 3.5) to[short] (10.25, 3.5);
	\draw (8.75, 3) to[short] (10.25, 3);
	\draw (8.75, 2.5) to[short] (10.25, 2.5);
	\draw[fill={rgb,255:red,255;green,255;blue,255}, line width=0.75pt] (9, 4) -- (9, 2) -- (10, 2.25) -- (10, 3.75) -- cycle node[anchor=center] at (9.5, 3){$\text{Bank}$};
\end{tikzpicture}

%% file: figures/mse.tikz
\begin{tikzpicture}[transform shape, scale=0.7]
	\draw (24, -16.02) to[american resistor, l_={$R_0$}] (22, -15.02);
	\node[fd inst amp, yscale=-1](vccs) at (25.56, -16.51){} node[anchor=center] at (vccs.text){$\beta$};
	\draw (22, -17.02) to[american resistor, l_={$R_0$}] (24, -16.02);
	\draw (24.37, -18) to[short] (24.37, -17);
	\draw (28, -17) to[short, i_<=$i_1^\mathrm{nudge}$] (26.75, -17);
	\draw (26.75, -16.02) to[short] (28, -15);
	\draw (24.37, -18) to[american controlled voltage source, l={${y_1^\mathrm{target}}$}, name=target] (24.37, -19);
	\draw (24, -16.02) to[short] (24.37, -16.02);
	\node[ocirc](N1) at (24, -16.02){} node[anchor=north] at (N1.south){$y_1$};
	\node[ocirc](N2) at (28, -17.02){} node[anchor=north] at (N2.south){$1'$};
	\node[ocirc](N3) at (28, -15.02){} node[anchor=south] at (N3.north){$1$};
	\node[ocirc](N4) at (22, -17.02){} node[anchor=north] at (N4.south){$1'$};
	\node[ocirc](N5) at (22, -15.02){} node[anchor=south] at (N5.north){$1$};
	\draw (24, -20.02) to[american resistor, l_={$R_0$}] (22, -19.02);
	\node[fd inst amp, yscale=-1](vccs) at (25.56, -20.51){} node[anchor=center] at (vccs.text){$\beta$};
	\draw (22, -21.02) to[american resistor, l_={$R_0$}] (24, -20.02);
	\draw (24.37, -22) to[short] (24.37, -21);
	\draw (28, -21.02) to[short, i_<=${i_2^\mathrm{nudge}}$] (26.75, -21);
	\draw (26.75, -20.02) to[short] (28, -19);
	\draw (24.37, -22) to[american controlled voltage source, l={${y_2^\mathrm{target}}$}, name=target] (24.37, -23);
	\draw (24, -20.02) to[short] (24.37, -20.02);
	\node[ocirc](N6) at (24, -20.02){} node[anchor=north] at (N6.south){$y_2$};
	\node[ocirc](N7) at (28, -21.02){} node[anchor=north] at (N7.south){$2'$};
	\node[ocirc](N8) at (28, -19.02){} node[anchor=south] at (N8.north){$2$};
	\node[ocirc](N9) at (22, -21.02){} node[anchor=north] at (N9.south){$2'$};
	\node[ocirc](N10) at (22, -19.02){} node[anchor=south] at (N10.north){$2$};
	\node[ground] at (24.37, -23){};
\end{tikzpicture}